\documentclass[letterpaper]{article} 
\usepackage{arxiv}  
\usepackage{times}  
\usepackage{helvet}  
\usepackage{courier}  
\usepackage[hyphens]{url}  
\usepackage{graphicx} 
\usepackage{natbib}  
\usepackage{caption} 
\usepackage{algorithm}
\usepackage{algorithmic}

\usepackage{newfloat}
\usepackage{listings}
\DeclareCaptionStyle{ruled}{labelfont=normalfont,labelsep=colon,strut=off} 
\floatstyle{ruled}
\newfloat{listing}{tb}{lst}{}
\floatname{listing}{Listing}
\usepackage{amsmath}
\usepackage{amsfonts}
\usepackage{booktabs}

\usepackage{pifont}
\usepackage{booktabs}
\usepackage{graphicx}
\usepackage[table]{xcolor}

\title{Self-Supervised Representation Learning with Joint Embedding Predictive Architecture for Automotive LiDAR Object Detection}

\author{
    Haoran Zhu\textsuperscript{\rm 1}\thanks{Main contributor.},
    Zhenyuan Dong\textsuperscript{\rm 1}\thanks{Equal contribution.},
    Kristi Topollai\textsuperscript{\rm 1}\footnotemark[2],
    Beiyao Sha\textsuperscript{\rm 1}\footnotemark[2],
    Anna Choromanska\textsuperscript{\rm 1}
}
\affiliations{
    \textsuperscript{\rm 1}Learning Systems Laboratory, Department of Electrical and Computer Engineering,\\
    New York University, 370 Jay Street, NY, USA\\
    \texttt{\{hz1922, zd2362, kt2664, bs5125, ac5455\}@nyu.edu}
}

\begin{document}

\maketitle

\begin{abstract}

Recently, self‑supervised representation learning relying on vast amounts of unlabeled data has been explored as a pre‑training method for autonomous driving. However, directly applying popular contrastive or generative methods to this problem is insufficient and may even lead to negative transfer. In this paper, we present \textbf{AD‑L‑JEPA}, a novel self‑supervised pre‑training framework with a joint embedding predictive architecture (JEPA) for automotive LiDAR object detection. Unlike existing methods, AD‑L‑JEPA is neither generative nor contrastive. Instead of explicitly generating masked regions, our method predicts Bird’s‑Eye‑View embeddings to capture the diverse nature of driving scenes. Furthermore, our approach eliminates the need to manually form contrastive pairs by employing explicit variance regularization to avoid representation collapse. Experimental results demonstrate consistent improvements on the LiDAR 3D object detection downstream task across the KITTI3D, Waymo, and ONCE datasets, while reducing GPU hours by $1.9\times$--$2.7\times$ and GPU memory by $2.8\times$--$4\times$ compared with the state-of-the-art method Occupancy-MAE. Notably, on the largest ONCE dataset, pre‑training on 100K frames yields a 1.61 mAP gain, better than in case of all the other methods pre‑trained on either 100K or 500K frames, and pre‑training on 500K frames yields a 2.98 mAP gain, better than in case of all the other methods pre‑trained on either 500K or 1M frames. AD‑L‑JEPA constitutes the first JEPA‑based pre‑training method for autonomous driving. It offers \textit{better quality}, \textit{faster}, and more \textit{GPU‑memory‑efficient} self‑supervised representation learning. The source code of AD-L-JEPA is ready to be released.
\end{abstract}


\begin{figure}[ht!]    \includegraphics[width=0.9\linewidth]{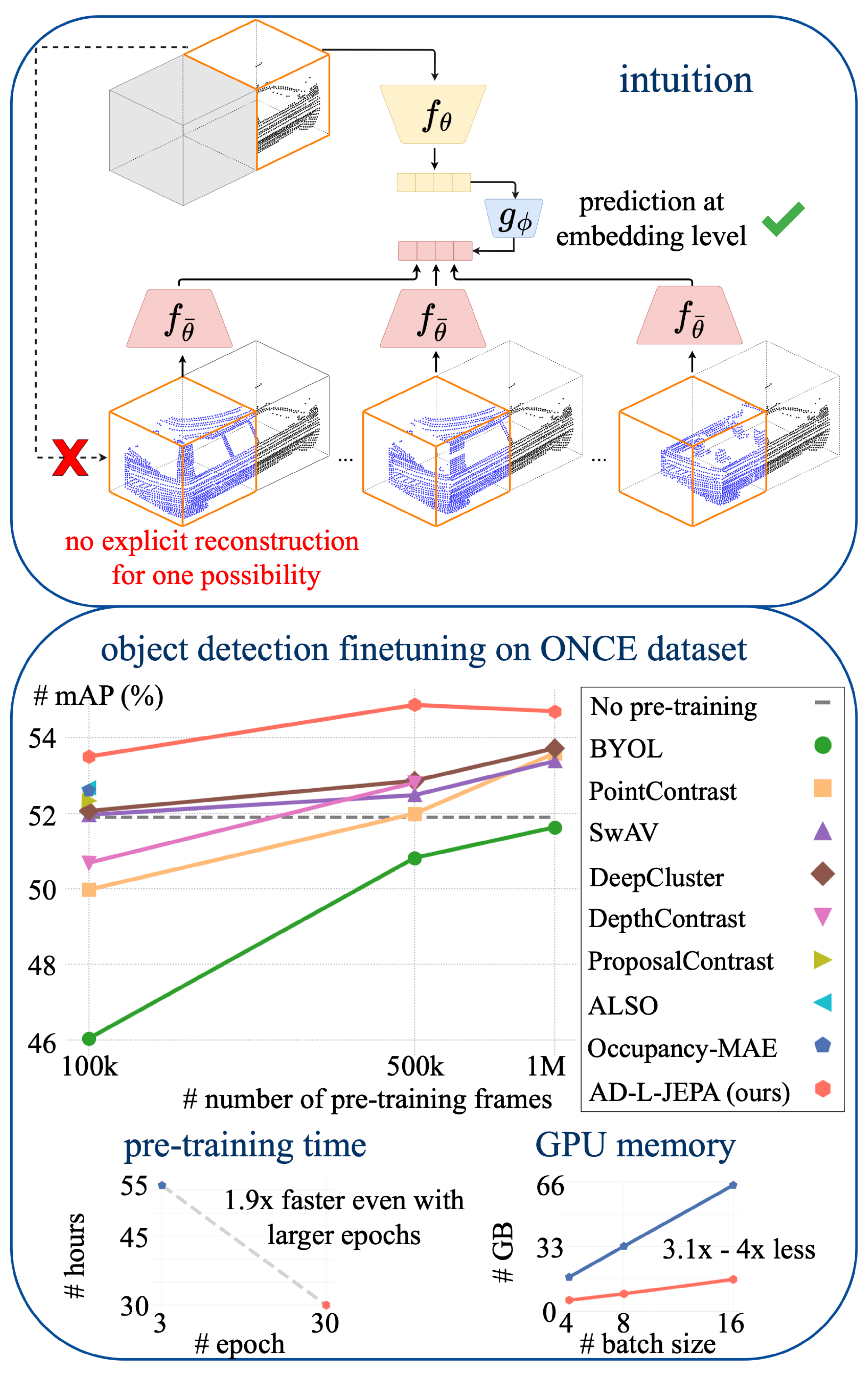}
    \caption{\textbf{AD‑L‑JEPA} predicts directly in embedding space instead of explicitly reconstructing masked point clouds, as in driving scenes these can correspond to multiple plausible point clouds sharing the same semantics (e.g., ``car rear'') and can be encoded into the same embedding. It significantly boosts downstream performance while reducing pre-training time and GPU memory usage.}
    \label{fig:intuition}
\end{figure}

\section{Introduction}
Unlike human drivers, current autonomous driving systems still require large amounts of labeled data for training. This supervised‑only paradigm is expensive due to labeling costs and limits the scalability of these systems. Recently, researchers have proposed a learning paradigm that relies on having a common backbone for different learning tasks that is pre-trained with self-supervised learning (SSL) across camera, LiDAR, and radar modalities~\cite{yang2024visual, min2023occupancy, zhu2024multi} without any labels and then fine-tuning the whole network with labeled data to adapt to specific downstream tasks. This approach requires less labeled data and improves generalization capability.

In SSL, the two most popular learning paradigms are contrastive methods~\cite{he2020momentum, chen2020simple} and generative methods~\cite{he2022masked}. However, directly applying these methods for pre‑training in autonomous driving is challenging and can even hurt downstream performance~\cite{mao2021one}. This stems from both the difficulty of defining meaningful contrastive pairs via data augmentation in driving scenarios that contain multiple objects, and the fact that explicit scene generation is time‑consuming and insufficient to capture semantic representations of diverse driving scenarios.

In this paper, we present AD-L-JEPA (aka \textbf{A}utonomous \textbf{D}riving with \textbf{L}iDAR data via a \textbf{J}oint \textbf{E}mbedding \textbf{P}redictive \textbf{A}rchitecture~\cite{lecun2022path}), a novel self-supervised pre-training framework for automotive LiDAR object detection that, as opposed to existing methods, is neither generative nor contrastive. Our method learns self‑supervised representations in Bird’s Eye View (BEV) space and predicts embeddings for spatially masked regions. It omits the need to create human‑crafted positive/negative pairs, as required by contrastive learning. Furthermore, rather than explicitly reconstructing unknown parts of the data as generative methods do, it predicts BEV embeddings instead.

AD‑L‑JEPA improves the quality of data representations, capturing high‑level semantics and better adapting to uncertainty across multiple plausible driving scenarios (toy illustration capturing that AD-L-JEPA can nicely handle data uncertainty is shown in Figure~\ref{fig:intuition}, where different car rears for masked regions are plausible given the visible car front), leading to \textit{better generalization}. It is also extremely \textit{efficient} and \textit{significantly reduces the pre-training time and GPU memory usage}. Extensive experiments show that AD-L-JEPA learns \textit{high-quality embeddings}, which result in \textit{a consistent boost in fine-tuning and transfer learning and offer superior label efficiency for downstream tasks}. A motivating illustration of AD‑L‑JEPA’s advantages is shown in Figure~\ref{fig:intuition}. Finally, AD‑L‑JEPA introduces a new approach to the existing family of self‑supervised pre‑training methods for autonomous driving, which compares favorably to the state‑of‑the‑art methods: Occupancy‑MAE~\cite{min2023occupancy} and ALSO~\cite{boulch2023also}. To the best of our knowledge, this is the first JEPA‑based pre‑training framework for autonomous driving.

\section{Related Work}
\subsection{Self-Supervised Learning}
Self-supervised learning~\cite{Balestriero2023ACO, he2020momentum, chen2020simple, grill2020bootstrap, chen2021exploring, caron2021emerging, zbontar2021barlow, bardes2022vicreg, he2022masked, assran2023self, bardes2024revisiting, assran2025v} aims at learning useful data representations from the unlabeled data. We focus here on SSL appraoches that are dedicated to autonomous driving with LiDAR point cloud data. Among the existing methods, contrastive techniques learn self-supervised representations by maximizing the feature similarity of manually created positive pairs (matching points) from different levels, such as point level~\cite{xie2020pointcontrast}, depth map level~\cite{zhang2021depthcontrast}, region clusters level~\cite{yin2022proposalcontrast}, spatial or temporal segments~\cite{liang2021gcc3d, huang2021spatio, nunes2022segcontrast, wu2023spatiotemporal, nunes2023temporal, yuan2024ad, wei2025t, hegde2025equivariant}, or Bird’s Eye View (BEV) features over time~\cite{sautier2024bevcontrast}, while minimizing the similarity of negative pairs (non-matched points), if any. Another family of approaches are the generative-based method, where the network is trained to reconstruct masked point clouds~\cite{min2023occupancy, lin2024bev, xu2023mv, wei2025t, abdelsamad2025multi} or scene surfaces~\cite{boulch2023also, agro2024uno}.

Besides contrastive and generative methods, another novel self-supervised learning paradigm offers joint-embedding predictive architecture (JEPA). It has recently been applied to image and video data modalities~\cite{lecun2022path, assran2023self, bardes2024revisiting, assran2025v} but has not yet been explored for autonomous driving scenarios. JEPA learns meaningful representations by predicting the embeddings of the unknown parts of data extracted by a target encoder given the known parts embeddings extracted by a context encoder. Unlike contrastive learning, it does not require complicated pre-processing to create positive pairs and does not suffer from the curse of dimensionality with negative pairs~\cite{lecun2022path}. Furthermore, it effectively captures the high uncertainty of the environment without explicitly generating reconstructions of unknown regions that is typically done by generative methods.

Representation collapse is a common phenomenon, when the network fails to learn meaningful representations. It can manifest in two ways: complete collapse, where all representations reduce to a constant vector and thus become useless, or dimensional collapse, where the dimensions of learned representations are highly correlated with each other and contain redundant information. Introducing contrastive learning, or applying regularization techniques~\cite{bardes2022vicreg} and moving average updates of the target encoder have been proven useful~\cite{grill2020bootstrap}. In our paper, we employ both regularization techniques and moving average updates of the target encoder to prevent collapse.

\subsection{LiDAR-Based 3D Object Detection}
We use 3D Object Detection as the downstream task to evaluate the effectiveness of our pre-training method for autonomous driving with LiDAR data. For most LiDAR-based 3D object detection downstream algorithms, a sparse 3D convolution encoder~\cite{graham2017submanifold} first extracts 3D voxel representations from a given point cloud scene. It then reshapes the image over the height dimension and applies 2D dense convolutions to generate BEV representations. Finally, either a single-stage~\cite{yan2018second} or a two-stage detection head~\cite{shi2019pointrcnn, yin2021center, shi2020pv, deng2021voxel} is attached for the LiDAR-based 3D object detection task. Following common settings in the literature, we focus on conducting self-supervised pre-training and evaluate the pre-trained feature quality also using the sparse 3D convolution encoder framework.

\begin{figure*}[ht!]
    \includegraphics[width=\linewidth]{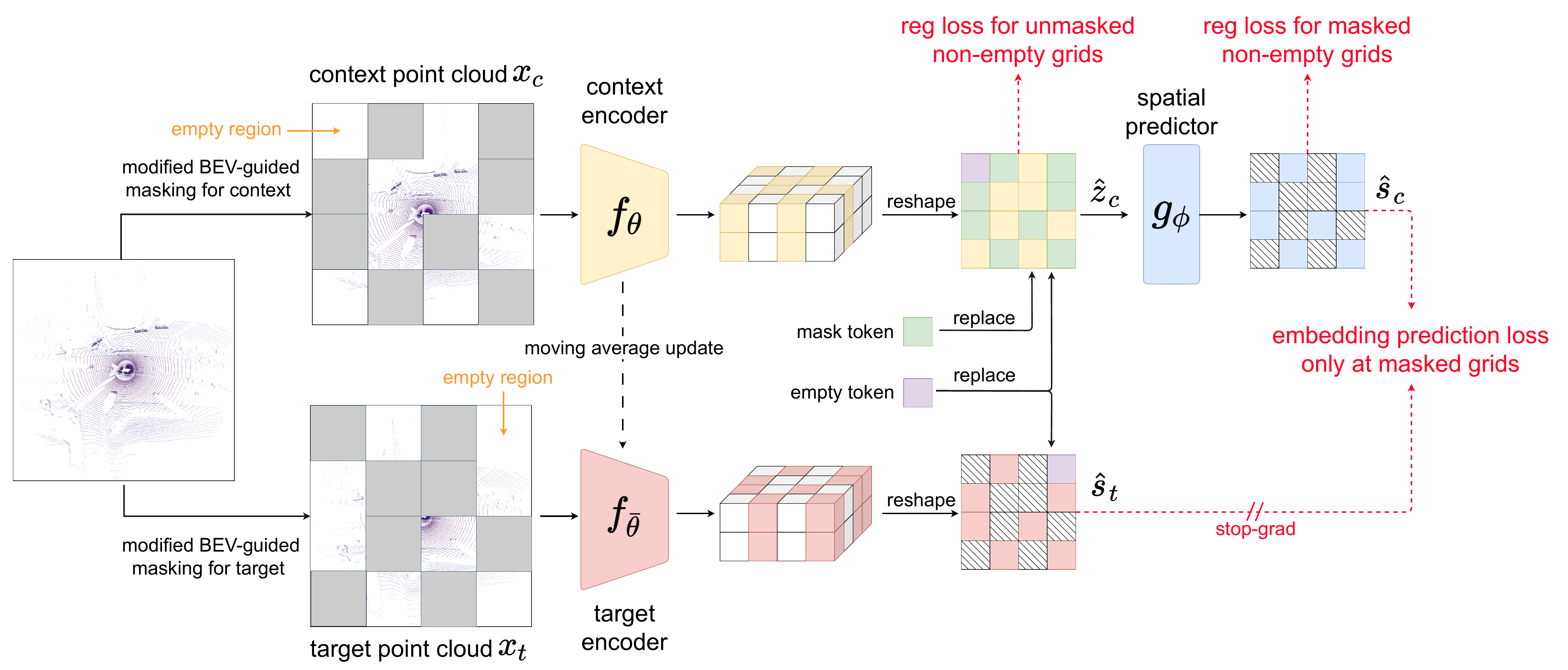}
    \caption{Overview of the AD-L-JEPA architecture: We introduce modified BEV-guided masking to mask the input point cloud in both empty and non-empty regions. The network predicts BEV embeddings at masked regions, leveraging variance regularization at non-empty regions following the output of the context encoder and the lightweight spatial predictor. It also employs a moving average update of the target encoder to learn diverse, high-level semantic representations.}
    \label{fig:architecture}
\end{figure*}

\section{Method}
The architecture of AD-L-JEPA is shown in Figure~\ref{fig:architecture}. The overarching intuition behind our framework is as follows: for the visible parts of the point cloud scene, the network is trained in a self-supervised manner to predict how the invisible parts should appear in the embedding space. This enables the learning of geometrically and semantically reasonable representations, as well as adapting to the high uncertainty nature of the autonomous driving scenes by avoiding the explicit reconstruction of the invisible parts of the data. In the following subsections we describe our design in detail.


\subsection{Modified BEV-Guided Masking}

\label{sec:masking}
\begin{figure}[ht!]
    \includegraphics[width=\linewidth]{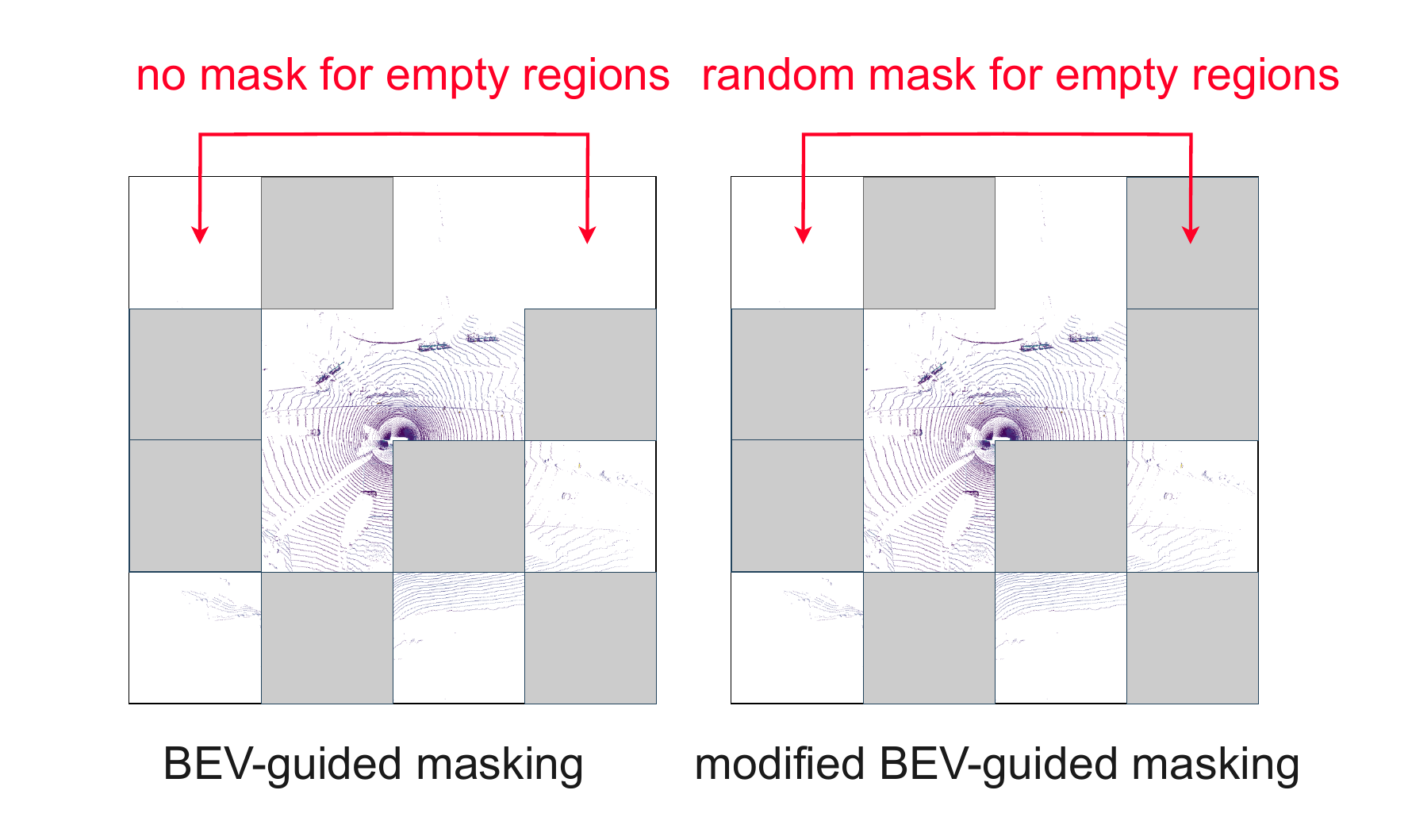}
    \caption{Comparison of original BEV-guided masking~\cite{lin2024bev} with our modified version that creates masks in both empty and non-empty regions in non-overlapping BEV grids.}
    \label{fig:masking}
\end{figure}

To learn effective representations in a self-supervised manner, masking is used to create invisible and visible regions. The network is then trained to predict embeddings of the invisible regions based on the visible ones. We have two design recipes for masking in autonomous driving scenarios: (1) masks are first created in the BEV embedding space and recursively upsampled to the input point cloud to identify points to be masked; (2) both empty and non-empty areas should be included in the visible and invisible regions created by the masks. These two criteria can be achieved by modifying the BEV-guided masking originally proposed in~\cite{lin2024bev} (see comparison in Figure~\ref{fig:masking}).

The original BEV-guided masking involves projecting points onto non-overlapping BEV grids, creating masks only in non-empty BEV grids, and reconstructing only in the masked non-empty regions. This assumes that the network already knows which BEV grid is empty, which stands in contrast with our goal to learn representations to predict all invisible regions. As opposed to this approach, we create masks for both empty and non-empty BEV grids, resulting in visible and invisible parts that contain both empty and non-empty regions. Consequently, the network is trained to predict the embeddings of all invisible parts, including those that contain empty regions, which enhances the encoder's representation power and helps to learn good quality representations. In our experiments, masks are applied to $50\%$ of non-empty BEV grids and $50\%$ of empty BEV grids. The input points sent to the encoder remain the same as in the original BEV-guided masking. However, the encoder's output BEV embeddings will later be replaced by mask tokens for both masked empty and non-empty grids, whereas in~\cite{lin2024bev}, only masked non-empty grids were replaced by mask tokens. We denote the unmasked point cloud as $\boldsymbol{x}_c=\{\boldsymbol{x}^1_c, \ldots, \boldsymbol{x}^N_c\}$ and the masked point cloud as $\boldsymbol{x}_t=\{\boldsymbol{x}^1_t, \ldots, \boldsymbol{x}^N_t\}$ in the multi-batch setting with a batch size of $N$. $\boldsymbol{x}_c$ and $\boldsymbol{x}_t$ are then sent to the context encoder and target encoder, respectively.

\subsection{Context Encoder \& Target encoder}
\label{sec:encoders}
The context encoder $f_\theta$ and target encoder $f_{\bar{\theta}}$ are backbones responsible for extracting context embeddings from the unmasked point cloud and target embeddings from the masked point cloud, respectively. The context encoder will later be used for fine-tuning on the downstream tasks after self-supervised representation learning. It receives input point cloud features and outputs embeddings in a downsampled 3D space $\in \mathbb{R}^{N \times H \times W \times D \times C}$, where $H$, $W$, and $D$ represent the length, width, and height dimensions of the 3D embedding, and $C$ is the embedding dimension. We obtain BEV embeddings by reshaping the 3D embeddings. The context BEV embedding is denoted as $\boldsymbol{z}_c = \textsf{reshape}(f_\theta(\boldsymbol{x}_c)) \in \mathbb{R}^{N \times H \times W \times D \cdot C}$, and the target BEV embeddings as $\boldsymbol{s}_t = \textsf{reshape}(f_{\bar{\theta}}(\boldsymbol{x}_t)) \in \mathbb{R}^{N \times H \times W \times D \cdot C}$.

\subsection{Learnable Empty Token and Mask Token}
\label{sec:add_tokens}
We introduce a learnable empty token ($\in \mathbb{R}^{D \cdot C}$) and a learnable mask token ($\in \mathbb{R}^{D \cdot C}$). By forwarding the unmasked point cloud to the context encoders and the masked point cloud to the target encoders, and after reshaping, we replace all unmasked empty grids in $\boldsymbol{z}_c$ and $\boldsymbol{s}_t$ with the learnable empty token. Simultaneously, we replace all masked grids in $\boldsymbol{z}_c$ with the learnable mask token. We then apply $L_2$ normalization to each BEV grid's embedding dimension. We denote the context embeddings and target embeddings after such a process by $\boldsymbol{\hat{z}}_c \in \mathbb{R}^{N \times H \times W \times D \cdot C}$ and $\boldsymbol{\hat{s}}_t \in \mathbb{R}^{N \times H \times W \times D \cdot C}$, respectively.

\subsection{Predictor}
\label{sec:predictor}
The predictor is a lightweight, three-layer convolutional network \( g_\phi \) that predicts target BEV embeddings from visible context BEV embeddings. We denote the predicted embedding, after the $L_2$ normalization is applied to each BEV grid's embedding dimension, as \( \boldsymbol{\hat{s}}_c = g_\phi(\boldsymbol{\hat{z}}_c) \), where \( \boldsymbol{\hat{s}}_c \) is in \( \mathbb{R}^{N \times H \times W \times D \cdot C} \) and has the same shape as the target BEV embeddings.

\subsection{Training}
\label{sec:training}
We pre-train the network in a self-supervised manner with two losses to ensure we learn high-quality, non-collapsed embeddings: a cosine similarity-based embedding prediction loss and a variance regularization loss.

For predicted embeddings $\boldsymbol{\hat{s}}_c$, we are primarily concerned with the embeddings of masked BEV grids, as we already know the embedding of unmasked BEV grids. Thus, we minimize the cosine similarity-based embedding prediction loss only for the masked grids. We also notice that objects in autonomous driving scenarios are distributed in a sparse manner; the number of BEV grids mapped with no points (empty grids) is significantly larger than the number of BEV grids mapped with points (non-empty grids). Therefore, we introduce the hyperparameters $\alpha_0=0.25$ and $\alpha_1=0.75$ to balance the loss. Overall, the prediction loss is:

\begin{equation}
\begin{aligned}
\mathcal{L_{\text{jepa}}}
&=\frac{\alpha_0}{\sum_{n=1}^N |P_n|} \sum_{n=1}^N \sum_{i=1}^{|P_n|}\left(1-\frac{\boldsymbol{\hat{s}}_c^n[i] \cdot \boldsymbol{\hat{s}}_t^n[i]}{\left\|\boldsymbol{\hat{s}}_c^n[i]\right\|\left\|\boldsymbol{\hat{s}}_t^n[i]\right\|}\right)\\
&+\frac{\alpha_1}{\sum_{n=1}^N |Q_n|} \sum_{n=1}^N \sum_{j=1}^{|Q_n|}\left(1-\frac{\boldsymbol{\hat{s}}_c^n[j] \cdot \boldsymbol{\hat{s}}_t^n[j]}{\left\|\boldsymbol{\hat{s}}_c^n[j]\right\|\left\|\boldsymbol{\hat{s}}_t^n[j]\right\|}\right),
\end{aligned}
\end{equation}
where \(P_n\) is the subset containing all indices of masked empty BEV grids, and \(Q_n\) is the subset containing all indices of masked non-empty BEV grids in the \(n\)-th input in a multi-batch setting. We also denote \(K_n\) as the subset containing all indices of masked non-empty BEV grids in the \(n\)-th input.

In order to avoid representation collapse, we apply an additional variance regularization loss proposed by~\cite{bardes2022vicreg}: 
\begin{equation}
    \mathcal{L_\text{reg}} = \beta_1 \sum_{n=1}^N   v\left(\boldsymbol{\hat{z}}_c^n[K_n]\right)+ \beta_2 \sum_{n=1}^N  v\left(\boldsymbol{\hat{s}}_c^n[Q_n]\right),
\end{equation}
where the function \( v(\cdot) \), which uses an arbitrary input 2D embedding matrix \( Y \in \mathbb{R}^{M \times C} \) recording an arbitrary number \( M \) of embeddings with a fixed embedding dimension \( C \), is defined in~\cite{bardes2022vicreg}:
\begin{equation}
    v(Y) = \frac{1}{C} \sum_{j=1}^{C} \max(0, \gamma - \sqrt{\operatorname{Var}(Y^j)+\epsilon})).
\end{equation} 

This loss ensures that the average variance across all embedding dimensions, for all unmasked non-empty grids' BEV embeddings after the context encoder $\boldsymbol{\hat{z}}_c^n[K_n] \in \mathbb{R}^{|K_n| \times C}$ , and all masked non-empty grids' BEV embeddings after the predictor $\boldsymbol{\hat{s}}_c^n[Q_n] \in \mathbb{R}^{|Q_n| \times C}$, is larger than some threshold $\gamma$. This approach prevents the learning of a meaningless constant embedding. Note that the regularization loss must be computed across each input in the multi-batch setting; otherwise, as we empirically observe, the network tends to learn another meaningless solution consisting of $N$ constant embeddings. These embeddings are significantly distinct from each other and are assigned to all non-empty grids in the $n$-th input, respectively. In this scenario, although the overall variance exceeds the threshold $\gamma$, each BEV embedding in the $n$-th input remains constant, which is meaningless. Averaging the loss across each input helps avoid this issue.

The overall self-supervised learning loss is:
\begin{equation}
\mathcal{L} = \lambda_{\text{jepa}} \mathcal{L_\text{jepa}} + \lambda_{\text{reg}} \mathcal{L_\text{reg}}
\label{eq:overall_loss}
\end{equation}

The parameters of the context encoder, the predictor, as well as the learnable mask token and the learnable empty token, are all updated by gradient descent using Figure~\ref{eq:overall_loss}. Meanwhile, the parameters of the target encoder are updated through a moving average of the context encoder's parameters, $\phi \leftarrow \eta \phi + (1 - \eta) \theta$, to further avoid representation collapse.

\section{Experiments}
To evaluate our pre-training method, We utilize the OpenPCDet framework \cite{openpcdet2020} to pre-process the datasets. We conduct experiments on three datasets of increasing scale: the small‑scale KITTI3D~\cite{geiger2012we} (7k frames), the medium‑scale Waymo~\cite{sun2020scalability} (30k frames for the 20\% subset and 150k frames for the full set), and the large‑scale ONCE~\cite{mao2021one} (100k, 500k, and 1M frames for the small, medium, and large splits). We then fine‑tune on a LiDAR 3D object detection task using different network architectures, including SECOND~\cite{yan2018second}, PV‑RCNN~\cite{shi2020pv}, and CenterPoint~\cite{yin2021center}.

We compare models trained from scratch and reproduce the most recent state-of-the-art (SOTA) self-supervised pre-training methods, Occupancy-MAE~\cite{min2023occupancy} and ALSO~\cite{boulch2023also} that are open-sourced and contain no running issues. Specifically, we reproduce Occupancy-MAE using their default settings to obtain the pre-trained weights and fine-tune the downstream object detector on the KITTI3D, Waymo, and ONCE 100k datasets. For ALSO, we directly fine-tune the released pre-trained encoder weights for the downstream object detector on the KITTI3D dataset. But for ALSO’s ONCE‑100k experiments, we observe that the released pre‑trained weights lead to significant negative transfer in three random runs, therefore we compare our results with those reported in their paper. The comparison methods use the same fine-tuning settings as ours.  For all fine-tuning experiments, we fine-tune the model through three independent runs in default settings and report the best performance, in the same manner as~\cite{sautier2024bevcontrast} reports their results. The reproduced results may vary from the original papers. Additionally, for the largest ONCE dataset, we include comparisons against the popular general self-supervised learning algorithms BYOL~\cite{grill2020bootstrap}, PointContrast~\cite{xie2020pointcontrast}, SwAV~\cite{caron2020unsupervised}, DeepCluster~\cite{caron2018deep}, and DepthContrast~\cite{zhang2021depthcontrast}, using results from ONCE’s official benchmark website. We also include a comparison to ProposalContrast~\cite{yin2022proposalcontrast}, using results reported in~\cite{min2023occupancy}.

Due to space constraints, additional details, such as dataset descriptions, hyperparameter settings, training protocols, visualizations, further ablation studies on masking ratios, predictor architecture, learnable empty/mask tokens, and variance regularization, as well as detailed comparisons of pre-training efficiency with other methods, etc., can be found in the supplementary material.

\subsection{Pre-training Efficiency}
Unlike Occupancy‑MAE, which uses computationally expensive dense 3D convolutions to reconstruct invisible regions, AD‑L‑JEPA employs a joint‑embedding predictive architecture at the BEV level and omits those layers, resulting in $2.8\times$–$3.4\times$ lower GPU memory usage and $2.7\times$ fewer GPU hours for pre‑training on the 20\% and 100\% splits of the Waymo dataset, and $3.1\times$–$4\times$ lower GPU memory usage and $1.9\times$ fewer GPU hours for pre‑training on the ONCE 100k split.

\subsection{Downstream Fine-tuning Performance}
We first conduct self-supervised pre-training on KITTI3D without any labels and then fine-tune the network on the labeled training data from the same KITTI3D dataset using the widely used SECOND~\cite{yan2018second} and PV-RCNN~\cite{shi2020pv} methods. As shown quantitatively in Table~\ref{table:kitti}, our self-supervised pre-training approach significantly outperforms the baseline that corresponds to training from scratch and also surpasses other competitor algorithms. 


\begin{table}[!ht]
\small
\setlength{\tabcolsep}{4.5pt}
\begin{tabular}{@{}l|c|c|c||cc@{}}
\toprule
 Method &  \multicolumn{1}{c}{Cars} &  \multicolumn{1}{c}{Ped.} &  \multicolumn{1}{c}{Cycl.} &  \multicolumn{1}{c}{Overall} & Diff.\\

\midrule
\multicolumn{5}{@{}l}{SECOND - $R_{40}$ metric}\\
No pre-training         & \textbf{81.99}  & 52.02 & 65.07 & 66.36 &         \\
Occupancy-MAE         & 81.15 & 50.36 & \textbf{69.74} & {67.08} &     {+0.72}       \\
ALSO& 81.48 &{52.50} & 65.95 & 66.64 &     {+0.28}       \\
AD-L-JEPA (ours) &  {81.68}  & \textbf{54.15} & {67.93} & \textbf{67.92} & \textbf{+1.56} \\

\midrule
\multicolumn{5}{@{}l}{SECOND - $R_{11}$ metric}\\

No pre-training             & \textbf{78.89} & 53.78 & 64.93 & 65.87 &             \\
Occupancy-MAE  &   78.15 & 52.10 & \textbf{69.39} & {66.55} & {+0.68} \\
ALSO& 78.37 & {53.94} & 66.06 & 66.12 &     {+0.25}       \\
AD-L-JEPA (ours) &   {78.51} & \textbf{54.86} & {67.94} & \textbf{67.10} & \textbf{+1.23} \\

\midrule
\multicolumn{5}{@{}l}{PV-RCNN - $R_{40}$ metric}\\
No pre-training         & {84.65}  & 56.19 & 72.19 & 71.01 &         \\
Occupancy-MAE        & 84.34 & {57.55} & 71.33 & 71.07 &     {+0.06}       \\
ALSO&  84.64 & 57.09 & \textbf{73.72} &  {71.82} &     {+0.81}       \\
AD-L-JEPA (ours) &  \textbf{85.07}  & \textbf{59.68} & {73.02} & \textbf{72.59} & \textbf{+1.58} \\

\midrule
\multicolumn{5}{@{}l}{PV-RCNN - $R_{11}$ metric}\\

No pre-training             & 83.38 & 57.01 & {72.65} & 71.01 &             \\
Occupancy-MAE   &   83.41 & {58.68} & 70.97 & 71.02 & {+0.01} \\
ALSO& {83.48} & 57.99 &  \textbf{72.89} & {71.45} &     {+0.44}       \\
AD-L-JEPA (ours) &   \textbf{83.74} & \textbf{59.91} & 72.19 & \textbf{71.95} & \textbf{+0.94} \\

\bottomrule
\end{tabular}
\caption{3D detection results on the KITTI3D validation set, reported with the AP (\%) metric. The models are pre-trained with and fine-tuned on the KITTI3D dataset.}
\label{table:kitti}
\end{table}


\begin{table}[!ht]
\small
\setlength{\tabcolsep}{4.5pt}
\begin{tabular}{@{}l|c|c|c||cc@{}}
\toprule
 Method &  \multicolumn{1}{c}{Veh.} &  \multicolumn{1}{c}{Ped.} &  \multicolumn{1}{c}{Cycl.} &  \multicolumn{1}{c}{Overall} & Diff.\\

\midrule
\multicolumn{5}{@{}l}{Centerpoint - AP metric}\\
No pre-training         &  63.28 & 63.95 & 66.77 & 64.67 &        \\
Occupancy-MAE, 20\%   &  63.20  & 64.20 & 67.20 &  64.87 & {+0.20} \\
Occupancy-MAE, 100\%  &  {63.53}  & \textbf{64.73} & {67.77} &  {65.34} & {+0.67} \\
AD-L-JEPA (ours), 20\% & 63.18 & 64.35 & 67.68 & 65.07 & {+0.40} \\
AD-L-JEPA (ours), 100\% & \textbf{63.58} & {64.58} & \textbf{68.07} & \textbf{65.41} & \textbf{+0.74} \\

\midrule
\multicolumn{5}{@{}l}{Centerpoint  - APH metric}\\
No pre-training         & 62.77  & 57.98 & 65.55 & 62.10 &         \\
Occupancy-MAE, 20\%   & 62.70  & 58.29  & 66.00 & 62.33 & {+0.23} \\
Occupancy-MAE, 100\%   &  {63.04}  & \textbf{58.81} & {66.55} &  {62.80} & {+0.70} \\
AD-L-JEPA (ours), 20\% & 62.68 & 58.39 & 66.48 & 62.51 & {+0.41} \\
AD-L-JEPA (ours), 100\% & \textbf{63.07} & {58.64} & \textbf{66.81} & \textbf{62.84} & \textbf{+0.74} \\

\bottomrule
\end{tabular}
\caption{3D detection results on the Waymo validation set with LEVEL\_2 difficulty defined in~\cite{sun2020scalability}, the AP (\%) and APH (\%) metric. The models are pre-trained with $20\%$ or $100\%$ of the Waymo dataset, as indicated, and are fine-tuned on the Waymo 20\% dataset.}
\label{table:waymo_detection}
\end{table}

\begin{table}[!ht]
\small
\setlength{\tabcolsep}{4.5pt}
\begin{tabular}{@{}l|c|c|c||cc@{}}
\toprule
 Method &  \multicolumn{1}{c}{Veh.} &  \multicolumn{1}{c}{Ped.} &  \multicolumn{1}{c}{Cycl.} &  \multicolumn{1}{c}{Overall} & Diff.\\
\midrule
No pre-training         & 71.19 & 26.44 & 58.04 & 51.89 &  \\
\midrule
\multicolumn{5}{@{}l}{SECOND, pre-trained with 100k frames}\\

BYOL                    & 68.02 & 19.50 & 50.61 & 46.04 & -5.85 \\
PointContrast           & 71.07 & 22.52 & 56.36 & 49.98 & -1.91 \\
SwAV                    & 72.71 & 25.13 & 58.05 & 51.96 & +0.07 \\
DeepCluster            & 73.19 & 24.00 & \textbf{58.99} & 52.06 & +0.17 \\
DepthContrast           & 71.88 & 23.57 & 56.63 & 50.69 & -1.20 \\
ProposalContrast        & 72.99 & 25.77 & 58.23 & 52.33 & +0.44 \\
ALSO                    & 71.73 & 28.16 & 58.13 & 52.68 & +0.79 \\
Occupancy-MAE          & \textbf{73.54} & 25.93 & 58.34 & 52.60 & +0.71 \\
AD-L-JEPA (ours)        & 73.18 & \textbf{29.19} & 58.14 & \textbf{53.50} & \textbf{+1.61} \\

\midrule
\multicolumn{5}{@{}l}{SECOND, pre-trained with 500k frames}\\
BYOL                    & 70.93  & 25.86 & 55.63 & 50.82 &  -1.07       \\
PointContrast           & 71.39  & 27.69 & 56.88 & 51.99 &  +0.10       \\
SwAV                   & 72.51  & 27.08 & 57.85 & 52.48 &  +0.59       \\

DepthContrast            & 71.92  & 29.01 & 57.51 & 52.81 &  +0.92       \\
DeepCluster             & 71.62  & 29.33 & 57.61 & 52.86 &  +0.97       \\

AD-L-JEPA (ours)        & \textbf{73.25}  & \textbf{31.91} & \textbf{59.47} & \textbf{54.87} &  \textbf{+2.98}       \\

\midrule
\multicolumn{5}{@{}l}{SECOND, pre-trained with 1M frames}\\
BYOL                    &  71.32 & 25.02 & 58.56 & 51.63 &  -0.26       \\
PointContrast           &  71.87 & 28.03 & \textbf{60.88} & 53.59 &  +1.70       \\
SwAV                    &  72.46 & 29.84 & 57.84 & 53.38 &  +1.49       \\
DeepCluster             &  72.89 & 30.32 & 57.94 & 53.72 &  +1.83       \\
AD-L-JEPA (ours),        & \textbf{73.01}  & \textbf{31.94} & 59.16 & \textbf{54.70} &  \textbf{+2.81}       \\

\bottomrule
\end{tabular}
\caption{3D detection results on the ONCE validation set with the AP (\%) metric. General self‑supervised learning leads to negligle or even negative gains; by contrast, AD‑L‑JEPA pre‑trained with 500k frames significantly outperforms models pre‑trained with 1M frames.}
\label{table:once_detection}
\end{table}

Next, in Table~\ref{table:waymo_detection}, we report the results of pre-training with 20\% or 100\% of the Waymo training data set, and fine-tuning with 20\% of the Waymo training data set with Centerpoint~\cite{yin2021center}. The overall Average Precision across all classes outperforms the baseline trained from scratch, and our reproduced results using Occupancy-MAE. 

Finally, Table~\ref{table:once_detection} reports the results of pre‑training on the largest ONCE dataset followed by fine‑tuning with SECOND. General self-supervised learning methods yield negligible or even negative gains, indicating the need for a domain specific design. AD-L-JEPA pre-trained on 100k frames outperforms all models trained on 100k or 500k frames and nearly matches the best model trained on 1M frames, and when pre-trained on 500k frames it surpasses every competitor, demonstrating scalability and superior self-supervised learning capability. Interestingly, AD-L-JEPA pre-trained on 1M frames, although significantly better than other methods, falls slightly behind AD-L-JEPA pre-trained on 500K frames, demonstrating that it is a method that is efficient in terms of data and learns high level useful representations with less unlabeled data. This small drop aligns with existing literature showing that increasing the number of unlabeled samples consistently boosts performance but saturates at a point \cite{goyal2019scaling}. Such saturation can be explained by the data redundancy of highly similar driving scenarios in the 1M frame setting and could potentially be mitigated by increasing the diversity of pre-training data \cite{al2024pretraining}.

\subsection{Transfer Learning and Label Efficiency}

\begin{table}[ht!]
\small
\setlength{\tabcolsep}{4.5pt}
\begin{tabular}{@{}l|c|c|c||cc@{}}
\toprule
 Method &  \multicolumn{1}{c}{Cars} &  \multicolumn{1}{c}{Ped.} &  \multicolumn{1}{c}{Cycl.} &  \multicolumn{1}{c}{Overall} & Diff.\\

\midrule
\multicolumn{5}{@{}l}{20\%, SECOND - $R_{40}$ metric}\\
No pre-training         & {79.11} & {44.36} & {62.55} & 62.01 &         \\
Occupancy-MAE        & 79.04  & 43.85 & \textbf{63.46} & {62.12} &     {+0.11}       \\
AD-L-JEPA (ours) &  \textbf{79.48}  & \textbf{48.48} & 61.92 & \textbf{63.30} & \textbf{+1.29} \\

\midrule
\multicolumn{5}{@{}l}{20\%, SECOND - $R_{11}$ metric}\\
No pre-training         & {77.84} & {45.78} & {62.46} & 62.03 &         \\
Occupancy-MAE         & 77.63 & 45.68 & \textbf{63.66} & {62.32} &     {+0.29}       \\
AD-L-JEPA (ours) &  \textbf{78.11}  & \textbf{49.71} & 62.18 & \textbf{63.33} & \textbf{+1.30} \\

\midrule
\multicolumn{5}{@{}l}{50\%, SECOND - $R_{40}$ metric}\\
No pre-training         & 81.05 & {48.75} & 62.83 & 64.21 &         \\
Occupancy-MAE         &  {81.20} & 48.17 & {64.09} & {64.49} &     {+0.28}       \\
AD-L-JEPA (ours) &  \textbf{81.55}  & \textbf{50.13} & \textbf{64.12} & \textbf{65.27} & \textbf{+1.06} \\

\midrule
\multicolumn{5}{@{}l}{50\%, SECOND - $R_{11}$ metric}\\
No pre-training         & 78.07 & {50.53} & 62.95 & 63.85 &         \\
Occupancy-MAE        &  {78.27} & 49.67 & \textbf{63.80}  &  {63.91} & {+0.06}       \\
AD-L-JEPA (ours) &  \textbf{78.38}  & \textbf{52.04} & {63.53} & \textbf{64.65} & \textbf{+0.80} \\

\midrule
\multicolumn{5}{@{}l}{100\%, SECOND - $R_{40}$ metric}\\
No pre-training         & \textbf{81.99}  & 52.02 & 65.07 & 66.36 &         \\
Occupancy-MAE       & 81.65 & 51.51 & 66.72 & 66.63 &     {+0.27}       \\
Occupancy-MAE\textsuperscript{\textdagger}         & 81.78 & 48.92 & {67.34} & 66.01 &     {-0.35}       \\
AD-L-JEPA (ours) &  {81.83}  & {52.41} & 66.73 & {66.99} & {+0.63} \\
AD-L-JEPA\textsuperscript{\textdagger} (ours) &  80.92  & \textbf{52.45} & \textbf{69.76} & \textbf{67.71} & \textbf{+1.35} \\

\midrule
\multicolumn{5}{@{}l}{100\%, SECOND- $R_{11}$ metric}\\
No pre-training             & \textbf{78.89} & {53.78} & 64.93 & 65.87 &             \\
Occupancy-MAE         & 78.47 & 52.60 &  
{67.23} & 66.10   &    {+0.23}       \\
Occupancy-MAE\textsuperscript{\textdagger}         & 78.61 & 49.49 & 66.84 & 64.98 &     {-0.89}       \\
AD-L-JEPA (ours) &  {78.65}  & 53.62 & 67.12 & {66.46} & {+0.59} \\
AD-L-JEPA\textsuperscript{\textdagger} (ours)&  77.90  & \textbf{54.01} & \textbf{69.30} & \textbf{67.07} & \textbf{+1.20} \\

\bottomrule
\end{tabular}
\caption{Transfer learning experiment: pre-training on Waymo ($20\%$ of the data set) and fine-tuning on KITTI3D. We report 3D detection results on the KITTI3D validation set, with the AP (\%) metric. Label efficiency is studied when fine-tuning with $20\%$, $50\%$, and $100\%$ of the data. \textdagger\ denotes that the first layer of the model is randomly initialized.}
\label{table:detection_best_over_3}
\end{table}

In this section, we report transfer learning from the Waymo 20\% split to KITTI3D. Each Waymo point has five attributes, whereas each KITTI3D point has only four. When fine‑tuning with an encoder pre‑trained directly on the Waymo 20\% split, we randomly initialize only the first layer (denoted \textsuperscript{\textdagger} at 100\% labels) and retain the remaining 3D encoder weights. Alternatively, we drop the fifth `elongation' feature, re‑pretrain on the first four attributes, and thus initialize every layer from pre‑trained weights. We also explore fine‑tune with 20\%, 50\%, and 100\% of KITTI labels in this setting. Table~\ref{table:detection_best_over_3} shows that AD‑L‑JEPA consistently outperforms baselines across different label efficiencies. Notably, $\text{AD-L-JEPA}^\text{{\textdagger}}$ model is strongest at 100\%, likely because retaining pre‑training on all original five features yields richer representations.

\begin{figure*}[!ht]
    \centering
    \includegraphics[width=\linewidth]{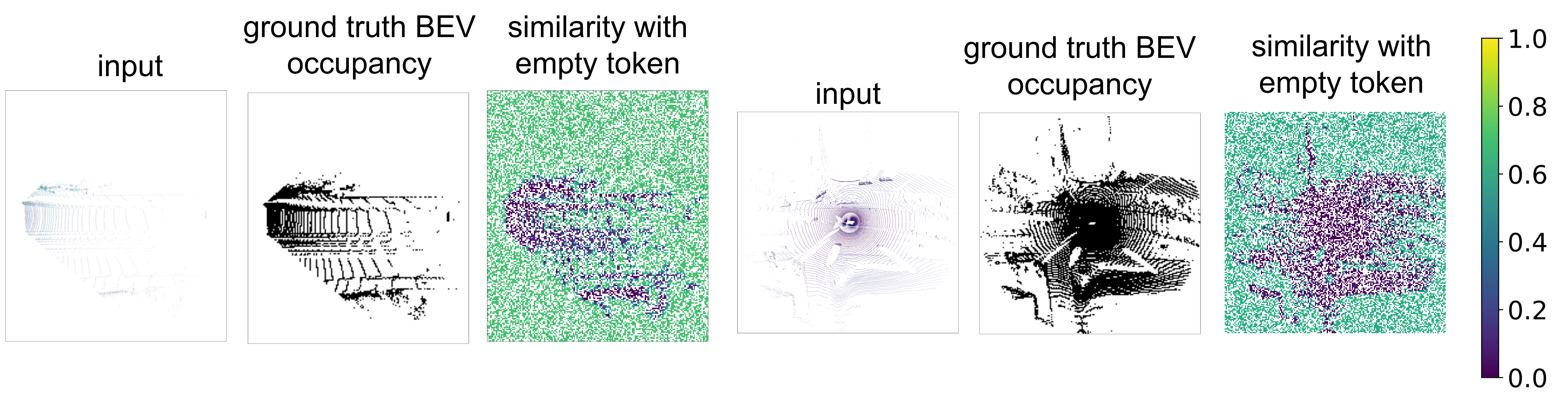}
    \caption{Masked region occupancy estimation evaluated by comparing BEV embeddings obtained by AD-L-JEPA with the learnable empty token via the cosine similarity. Unmasked regions are ignored and the cosine similarity in this case is represented in white color.}
    \label{fig:occupancy_estimation}
\end{figure*}

\begin{figure*}[!ht]
    \centering
    \includegraphics[width=\linewidth]{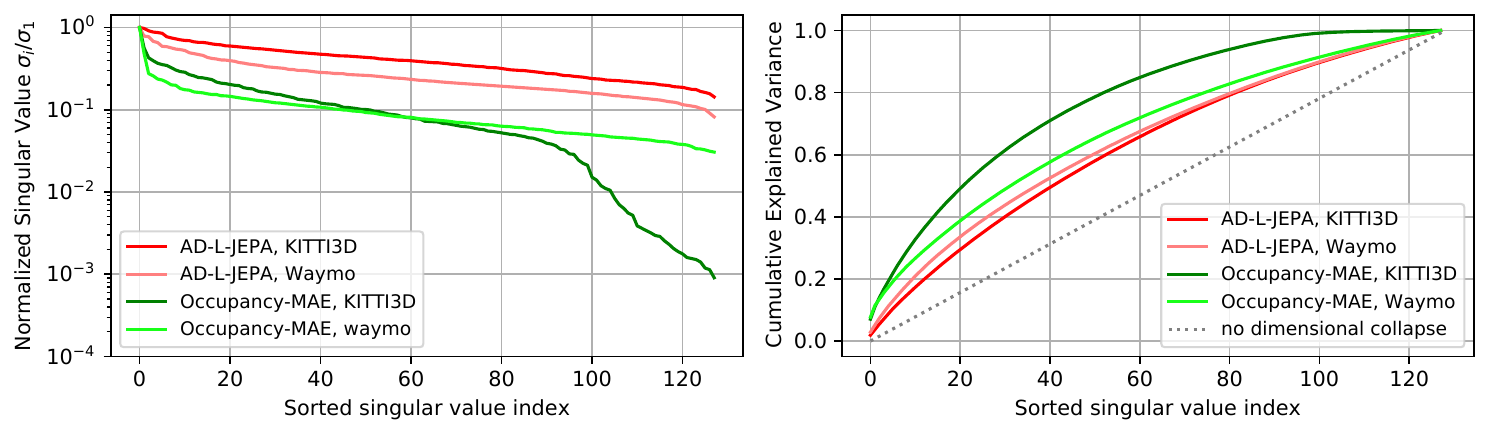}
    \caption{Sorted normalized singular values and the corresponding cumulative explained variance, obtained by singular value decomposition of pre-trained BEV embeddings. Embeddings are obtained either with AD-L-JEPA or Occupancy-MAE.}
    \label{fig:svd_analysis}
\end{figure*}
\subsection{Other Evaluations}

We also interpret the representation capability of AD-L-JEPA by the following method:

\subsubsection{Occupancy Estimation}
In Figure~\ref{fig:occupancy_estimation}, we  estimate the occupancy of masked regions in the input point cloud at a downsampled BEV embedding resolution by comparing each grid's BEV embeddings, outputted by our pre-trained model, against the learnable empty token using cosine similarity. The cosine similarity assigns high similarity to empty grids and low similarity to non-empty grids. We then compare the cosine similarity map with ground truth BEV occupancy side-by-side. Note that the similarity assigned to empty grids by our method is approximately $0.7$, not $1$, showcasing the model's ability to represent the complex and highly uncertain nature of autonomous driving scenes.

\subsubsection{Singular Value Decomposition Analysis}
We follow the methodology described in~\cite{li2022understanding} to conduct singular value decomposition of non-empty BEV embeddings outputted by encoders pre-trained with AD-L-JEPA and Occupancy-MAE. This analysis assesses the level of dimensional collapse in pre-trained embeddings. Figure~\ref{fig:svd_analysis} shows that the normalized sorted singular values for AD-L-JEPA are more evenly distributed, while the corresponding cumulative explained variance increases more slowly, indicating less dimensional collapse and less redundant information in the embeddings.

\section{Conclusions}
In this paper, we propose AD-L-JEPA, the first joint-embedding predictive architecture for self-supervised representation learning of autonomous-driving LiDAR data. It learns useful representations without any labeled data by predicting masked regions in the BEV embedding space. It neither requires manual creation of positive/negative pairs for contrastive learning nor explicitly reconstructs the complex, high-uncertainty driving scenes. Extensive experiments show that AD-L-JEPA is a more efficient pre-training method for automotive LiDAR-based object detection than state-of-the-art methods, learning richer representations that generalize better to downstream tasks while requiring significantly fewer GPU pre-training hours and less GPU memory. For future work, we plan to extend AD-L-JEPA to leverage temporal dynamics and to incorporate action-conditioned self-supervised representation learning in autonomous driving scenarios.

\newpage

\bibliography{files/LaTeX/references}

\begin{thebibliography}{48}
\providecommand{\natexlab}[1]{#1}

\bibitem[{Abdelsamad et~al.(2025)Abdelsamad, Ulrich, Gl{\"a}ser, and Valada}]{abdelsamad2025multi}
Abdelsamad, M.; Ulrich, M.; Gl{\"a}ser, C.; and Valada, A. 2025.
\newblock Multi-Scale Neighborhood Occupancy Masked Autoencoder for Self-Supervised Learning in LiDAR Point Clouds.
\newblock In \emph{Proceedings of the Computer Vision and Pattern Recognition Conference}, 22234--22243.

\bibitem[{Agro et~al.(2024)Agro, Sykora, Casas, Gilles, and Urtasun}]{agro2024uno}
Agro, B.; Sykora, Q.; Casas, S.; Gilles, T.; and Urtasun, R. 2024.
\newblock UnO: Unsupervised Occupancy Fields for Perception and Forecasting.
\newblock In \emph{Proceedings of the IEEE/CVF Conference on Computer Vision and Pattern Recognition}, 14487--14496.

\bibitem[{Al~Kader~Hammoud et~al.(2024)Al~Kader~Hammoud, Das, Pizzati, Torr, Bibi, and Ghanem}]{al2024pretraining}
Al~Kader~Hammoud, H.~A.; Das, T.; Pizzati, F.; Torr, P.~H.; Bibi, A.; and Ghanem, B. 2024.
\newblock On pretraining data diversity for self-supervised learning.
\newblock In \emph{European Conference on Computer Vision}, 54--71. Springer.

\bibitem[{Assran et~al.(2025)Assran, Bardes, Fan, Garrido, Howes, Muckley, Rizvi, Roberts, Sinha, Zholus et~al.}]{assran2025v}
Assran, M.; Bardes, A.; Fan, D.; Garrido, Q.; Howes, R.; Muckley, M.; Rizvi, A.; Roberts, C.; Sinha, K.; Zholus, A.; et~al. 2025.
\newblock V-jepa 2: Self-supervised video models enable understanding, prediction and planning.
\newblock \emph{arXiv preprint arXiv:2506.09985}.

\bibitem[{Assran et~al.(2023)Assran, Duval, Misra, Bojanowski, Vincent, Rabbat, LeCun, and Ballas}]{assran2023self}
Assran, M.; Duval, Q.; Misra, I.; Bojanowski, P.; Vincent, P.; Rabbat, M.; LeCun, Y.; and Ballas, N. 2023.
\newblock Self-supervised learning from images with a joint-embedding predictive architecture.
\newblock In \emph{Proceedings of the IEEE/CVF Conference on Computer Vision and Pattern Recognition}, 15619--15629.

\bibitem[{Balestriero et~al.(2023)Balestriero, Ibrahim, Sobal, Morcos, Shekhar, Goldstein, Bordes, Bardes, Mialon, Tian, Schwarzschild, Wilson, Geiping, Garrido, Fernandez, Bar, Pirsiavash, LeCun, and Goldblum}]{Balestriero2023ACO}
Balestriero, R.; Ibrahim, M.; Sobal, V.; Morcos, A.~S.; Shekhar, S.; Goldstein, T.; Bordes, F.; Bardes, A.; Mialon, G.; Tian, Y.; Schwarzschild, A.; Wilson, A.~G.; Geiping, J.; Garrido, Q.; Fernandez, P.; Bar, A.; Pirsiavash, H.; LeCun, Y.; and Goldblum, M. 2023.
\newblock A Cookbook of Self-Supervised Learning.
\newblock \emph{ArXiv}, abs/2304.12210.

\bibitem[{Bardes et~al.(2024)Bardes, Garrido, Ponce, Chen, Rabbat, LeCun, Assran, and Ballas}]{bardes2024revisiting}
Bardes, A.; Garrido, Q.; Ponce, J.; Chen, X.; Rabbat, M.; LeCun, Y.; Assran, M.; and Ballas, N. 2024.
\newblock Revisiting feature prediction for learning visual representations from video.
\newblock \emph{arXiv preprint arXiv:2404.08471}.

\bibitem[{Bardes, Ponce, and LeCun(2022)}]{bardes2022vicreg}
Bardes, A.; Ponce, J.; and LeCun, Y. 2022.
\newblock VICReg: Variance-Invariance-Covariance Regularization For Self-Supervised Learning.
\newblock In \emph{ICLR}.

\bibitem[{Boulch et~al.(2023)Boulch, Sautier, Michele, Puy, and Marlet}]{boulch2023also}
Boulch, A.; Sautier, C.; Michele, B.; Puy, G.; and Marlet, R. 2023.
\newblock Also: Automotive lidar self-supervision by occupancy estimation.
\newblock In \emph{Proceedings of the IEEE/CVF Conference on Computer Vision and Pattern Recognition}, 13455--13465.

\bibitem[{Caron et~al.(2018)Caron, Bojanowski, Joulin, and Douze}]{caron2018deep}
Caron, M.; Bojanowski, P.; Joulin, A.; and Douze, M. 2018.
\newblock Deep clustering for unsupervised learning of visual features.
\newblock In \emph{Proceedings of the European conference on computer vision (ECCV)}, 132--149.

\bibitem[{Caron et~al.(2020)Caron, Misra, Mairal, Goyal, Bojanowski, and Joulin}]{caron2020unsupervised}
Caron, M.; Misra, I.; Mairal, J.; Goyal, P.; Bojanowski, P.; and Joulin, A. 2020.
\newblock Unsupervised learning of visual features by contrasting cluster assignments.
\newblock \emph{Advances in neural information processing systems}, 33: 9912--9924.

\bibitem[{Caron et~al.(2021)Caron, Touvron, Misra, J{\'e}gou, Mairal, Bojanowski, and Joulin}]{caron2021emerging}
Caron, M.; Touvron, H.; Misra, I.; J{\'e}gou, H.; Mairal, J.; Bojanowski, P.; and Joulin, A. 2021.
\newblock Emerging properties in self-supervised vision transformers.
\newblock In \emph{Proceedings of the IEEE/CVF international conference on computer vision}, 9650--9660.

\bibitem[{Chen et~al.(2020)Chen, Kornblith, Norouzi, and Hinton}]{chen2020simple}
Chen, T.; Kornblith, S.; Norouzi, M.; and Hinton, G. 2020.
\newblock A simple framework for contrastive learning of visual representations.
\newblock In \emph{International conference on machine learning}, 1597--1607. PMLR.

\bibitem[{Chen and He(2021)}]{chen2021exploring}
Chen, X.; and He, K. 2021.
\newblock Exploring simple siamese representation learning.
\newblock In \emph{Proceedings of the IEEE/CVF conference on computer vision and pattern recognition}, 15750--15758.

\bibitem[{Deng et~al.(2021)Deng, Shi, Li, Zhou, Zhang, and Li}]{deng2021voxel}
Deng, J.; Shi, S.; Li, P.; Zhou, W.; Zhang, Y.; and Li, H. 2021.
\newblock Voxel r-cnn: Towards high performance voxel-based 3d object detection.
\newblock In \emph{Proceedings of the AAAI conference on artificial intelligence}.

\bibitem[{Geiger, Lenz, and Urtasun(2012)}]{geiger2012we}
Geiger, A.; Lenz, P.; and Urtasun, R. 2012.
\newblock Are we ready for autonomous driving? the kitti vision benchmark suite.
\newblock In \emph{2012 IEEE conference on computer vision and pattern recognition}, 3354--3361. IEEE.

\bibitem[{Goyal et~al.(2019)Goyal, Mahajan, Gupta, and Misra}]{goyal2019scaling}
Goyal, P.; Mahajan, D.; Gupta, A.; and Misra, I. 2019.
\newblock Scaling and benchmarking self-supervised visual representation learning.
\newblock In \emph{Proceedings of the ieee/cvf International Conference on computer vision}, 6391--6400.

\bibitem[{Graham and Van~der Maaten(2017)}]{graham2017submanifold}
Graham, B.; and Van~der Maaten, L. 2017.
\newblock Submanifold sparse convolutional networks.
\newblock \emph{arXiv preprint arXiv:1706.01307}.

\bibitem[{Grill et~al.(2020)Grill, Strub, Altch{\'e}, Tallec, Richemond, Buchatskaya, Doersch, Avila~Pires, Guo, Gheshlaghi~Azar et~al.}]{grill2020bootstrap}
Grill, J.-B.; Strub, F.; Altch{\'e}, F.; Tallec, C.; Richemond, P.; Buchatskaya, E.; Doersch, C.; Avila~Pires, B.; Guo, Z.; Gheshlaghi~Azar, M.; et~al. 2020.
\newblock Bootstrap your own latent-a new approach to self-supervised learning.
\newblock \emph{Advances in neural information processing systems}, 33: 21271--21284.

\bibitem[{He et~al.(2022)He, Chen, Xie, Li, Doll{\'a}r, and Girshick}]{he2022masked}
He, K.; Chen, X.; Xie, S.; Li, Y.; Doll{\'a}r, P.; and Girshick, R. 2022.
\newblock Masked autoencoders are scalable vision learners.
\newblock In \emph{Proceedings of the IEEE/CVF Conference on Computer Vision and Pattern Recognition}, 16000--16009.

\bibitem[{He et~al.(2020)He, Fan, Wu, Xie, and Girshick}]{he2020momentum}
He, K.; Fan, H.; Wu, Y.; Xie, S.; and Girshick, R. 2020.
\newblock Momentum contrast for unsupervised visual representation learning.
\newblock In \emph{Proceedings of the IEEE/CVF conference on computer vision and pattern recognition}, 9729--9738.

\bibitem[{Hegde et~al.(2025)Hegde, Lohit, Peng, Jones, and Patel}]{hegde2025equivariant}
Hegde, D.; Lohit, S.; Peng, K.-C.; Jones, M.~J.; and Patel, V.~M. 2025.
\newblock Equivariant Spatio-Temporal Self-Supervision for LiDAR Object Detection.
\newblock In \emph{European Conference on Computer Vision}, 475--491. Springer.

\bibitem[{Huang et~al.(2021)Huang, Xie, Zhu, and Zhu}]{huang2021spatio}
Huang, S.; Xie, Y.; Zhu, S.-C.; and Zhu, Y. 2021.
\newblock Spatio-temporal self-supervised representation learning for 3d point clouds.
\newblock In \emph{Proceedings of the IEEE/CVF International Conference on Computer Vision}, 6535--6545.

\bibitem[{LeCun(2022)}]{lecun2022path}
LeCun, Y. 2022.
\newblock A path towards autonomous machine intelligence version 0.9. 2, 2022-06-27.
\newblock \emph{Open Review}, 62.

\bibitem[{Li, Efros, and Pathak(2022)}]{li2022understanding}
Li, A.~C.; Efros, A.~A.; and Pathak, D. 2022.
\newblock Understanding collapse in non-contrastive siamese representation learning.
\newblock In \emph{European Conference on Computer Vision}, 490--505. Springer.

\bibitem[{Liang et~al.(2021)Liang, Jiang, Feng, Chen, Xu, Liang, Zhang, Li, and Van~Gool}]{liang2021gcc3d}
Liang, H.; Jiang, C.; Feng, D.; Chen, X.; Xu, H.; Liang, X.; Zhang, W.; Li, Z.; and Van~Gool, L. 2021.
\newblock Exploring geometry-aware contrast and clustering harmonization for self-supervised 3d object detection.
\newblock In \emph{Proceedings of the IEEE/CVF International Conference on Computer Vision}, 3293--3302.

\bibitem[{Lin et~al.(2024)Lin, Wang, Qi, Dong, and Yang}]{lin2024bev}
Lin, Z.; Wang, Y.; Qi, S.; Dong, N.; and Yang, M.-H. 2024.
\newblock BEV-MAE: Bird’s Eye View Masked Autoencoders for Point Cloud Pre-training in Autonomous Driving Scenarios.
\newblock In \emph{Proceedings of the AAAI Conference on Artificial Intelligence}, volume~38, 3531--3539.

\bibitem[{Mao et~al.(2021)Mao, Niu, Jiang, Liang, Chen, Liang, Li, Ye, Zhang, Li et~al.}]{mao2021one}
Mao, J.; Niu, M.; Jiang, C.; Liang, H.; Chen, J.; Liang, X.; Li, Y.; Ye, C.; Zhang, W.; Li, Z.; et~al. 2021.
\newblock One million scenes for autonomous driving: Once dataset.
\newblock \emph{arXiv preprint arXiv:2106.11037}.

\bibitem[{Min et~al.(2023)Min, Xiao, Zhao, Nie, and Dai}]{min2023occupancy}
Min, C.; Xiao, L.; Zhao, D.; Nie, Y.; and Dai, B. 2023.
\newblock Occupancy-mae: Self-supervised pre-training large-scale lidar point clouds with masked occupancy autoencoders.
\newblock \emph{IEEE Transactions on Intelligent Vehicles}.

\bibitem[{Nunes et~al.(2022)Nunes, Marcuzzi, Chen, Behley, and Stachniss}]{nunes2022segcontrast}
Nunes, L.; Marcuzzi, R.; Chen, X.; Behley, J.; and Stachniss, C. 2022.
\newblock SegContrast: 3D point cloud feature representation learning through self-supervised segment discrimination.
\newblock \emph{IEEE Robotics and Automation Letters}, 7(2): 2116--2123.

\bibitem[{Nunes et~al.(2023)Nunes, Wiesmann, Marcuzzi, Chen, Behley, and Stachniss}]{nunes2023temporal}
Nunes, L.; Wiesmann, L.; Marcuzzi, R.; Chen, X.; Behley, J.; and Stachniss, C. 2023.
\newblock Temporal consistent 3D lidar representation learning for semantic perception in autonomous driving.
\newblock In \emph{Proceedings of the IEEE/CVF Conference on Computer Vision and Pattern Recognition}, 5217--5228.

\bibitem[{Sautier et~al.(2024)Sautier, Puy, Boulch, Marlet, and Lepetit}]{sautier2024bevcontrast}
Sautier, C.; Puy, G.; Boulch, A.; Marlet, R.; and Lepetit, V. 2024.
\newblock Bevcontrast: Self-supervision in bev space for automotive lidar point clouds.
\newblock In \emph{2024 International Conference on 3D Vision (3DV)}, 559--568. IEEE.

\bibitem[{Shi et~al.(2020)Shi, Guo, Jiang, Wang, Shi, Wang, and Li}]{shi2020pv}
Shi, S.; Guo, C.; Jiang, L.; Wang, Z.; Shi, J.; Wang, X.; and Li, H. 2020.
\newblock Pv-rcnn: Point-voxel feature set abstraction for 3d object detection.
\newblock In \emph{Proceedings of the IEEE/CVF conference on computer vision and pattern recognition}, 10529--10538.

\bibitem[{Shi, Wang, and Li(2019)}]{shi2019pointrcnn}
Shi, S.; Wang, X.; and Li, H. 2019.
\newblock Pointrcnn: 3d object proposal generation and detection from point cloud.
\newblock In \emph{Proceedings of the IEEE/CVF conference on computer vision and pattern recognition}, 770--779.

\bibitem[{Sun et~al.(2020)Sun, Kretzschmar, Dotiwalla, Chouard, Patnaik, Tsui, Guo, Zhou, Chai, Caine et~al.}]{sun2020scalability}
Sun, P.; Kretzschmar, H.; Dotiwalla, X.; Chouard, A.; Patnaik, V.; Tsui, P.; Guo, J.; Zhou, Y.; Chai, Y.; Caine, B.; et~al. 2020.
\newblock Scalability in perception for autonomous driving: Waymo open dataset.
\newblock In \emph{Proceedings of the IEEE/CVF conference on computer vision and pattern recognition}, 2446--2454.

\bibitem[{Team(2020)}]{openpcdet2020}
Team, O.~D. 2020.
\newblock OpenPCDet: An Open-source Toolbox for 3D Object Detection from Point Clouds.
\newblock \url{https://github.com/open-mmlab/OpenPCDet}.

\bibitem[{Wei et~al.(2025)Wei, Nejadasl, Gevers, and Oswald}]{wei2025t}
Wei, W.; Nejadasl, F.~K.; Gevers, T.; and Oswald, M.~R. 2025.
\newblock T-MAE: temporal masked autoencoders for point cloud representation learning.
\newblock In \emph{European Conference on Computer Vision}, 178--195. Springer.

\bibitem[{Wu et~al.(2023)Wu, Zhang, Ke, S{\"u}sstrunk, and Salzmann}]{wu2023spatiotemporal}
Wu, Y.; Zhang, T.; Ke, W.; S{\"u}sstrunk, S.; and Salzmann, M. 2023.
\newblock Spatiotemporal self-supervised learning for point clouds in the wild.
\newblock In \emph{Proceedings of the IEEE/CVF Conference on Computer Vision and Pattern Recognition}, 5251--5260.

\bibitem[{Xie et~al.(2020)Xie, Gu, Guo, Qi, Guibas, and Litany}]{xie2020pointcontrast}
Xie, S.; Gu, J.; Guo, D.; Qi, C.~R.; Guibas, L.; and Litany, O. 2020.
\newblock Pointcontrast: Unsupervised pre-training for 3d point cloud understanding.
\newblock In \emph{Computer Vision--ECCV 2020: 16th European Conference, Glasgow, UK, August 23--28, 2020, Proceedings, Part III 16}, 574--591. Springer.

\bibitem[{Xu et~al.(2023)Xu, Wang, Zhang, Chen, Cao, Pang, and Lin}]{xu2023mv}
Xu, R.; Wang, T.; Zhang, W.; Chen, R.; Cao, J.; Pang, J.; and Lin, D. 2023.
\newblock Mv-jar: Masked voxel jigsaw and reconstruction for lidar-based self-supervised pre-training.
\newblock In \emph{Proceedings of the IEEE/CVF Conference on Computer Vision and Pattern Recognition}, 13445--13454.

\bibitem[{Yan, Mao, and Li(2018)}]{yan2018second}
Yan, Y.; Mao, Y.; and Li, B. 2018.
\newblock Second: Sparsely embedded convolutional detection.
\newblock \emph{Sensors}, 18(10): 3337.

\bibitem[{Yang et~al.(2024)Yang, Chen, Sun, and Li}]{yang2024visual}
Yang, Z.; Chen, L.; Sun, Y.; and Li, H. 2024.
\newblock Visual point cloud forecasting enables scalable autonomous driving.
\newblock In \emph{Proceedings of the IEEE/CVF Conference on Computer Vision and Pattern Recognition}, 14673--14684.

\bibitem[{Yin et~al.(2022)Yin, Zhou, Zhang, Fang, Xu, Shen, and Wang}]{yin2022proposalcontrast}
Yin, J.; Zhou, D.; Zhang, L.; Fang, J.; Xu, C.-Z.; Shen, J.; and Wang, W. 2022.
\newblock Proposalcontrast: Unsupervised pre-training for lidar-based 3d object detection.
\newblock In \emph{European conference on computer vision}, 17--33. Springer.

\bibitem[{Yin, Zhou, and Krahenbuhl(2021)}]{yin2021center}
Yin, T.; Zhou, X.; and Krahenbuhl, P. 2021.
\newblock Center-based 3d object detection and tracking.
\newblock In \emph{Proceedings of the IEEE/CVF conference on computer vision and pattern recognition}, 11784--11793.

\bibitem[{Yuan et~al.(2024)Yuan, Zhang, Yan, Shi, Chen, Li, and Qiao}]{yuan2024ad}
Yuan, J.; Zhang, B.; Yan, X.; Shi, B.; Chen, T.; Li, Y.; and Qiao, Y. 2024.
\newblock Ad-pt: Autonomous driving pre-training with large-scale point cloud dataset.
\newblock \emph{Advances in Neural Information Processing Systems}, 36.

\bibitem[{Zbontar et~al.(2021)Zbontar, Jing, Misra, LeCun, and Deny}]{zbontar2021barlow}
Zbontar, J.; Jing, L.; Misra, I.; LeCun, Y.; and Deny, S. 2021.
\newblock Barlow twins: Self-supervised learning via redundancy reduction.
\newblock In \emph{International Conference on Machine Learning}, 12310--12320. PMLR.

\bibitem[{Zhang et~al.(2021)Zhang, Girdhar, Joulin, and Misra}]{zhang2021depthcontrast}
Zhang, Z.; Girdhar, R.; Joulin, A.; and Misra, I. 2021.
\newblock Self-supervised pretraining of 3d features on any point-cloud.
\newblock In \emph{Proceedings of the IEEE/CVF International Conference on Computer Vision}, 10252--10263.

\bibitem[{Zhu et~al.(2024)Zhu, He, Choromanska, Ravindran, Shi, and Chen}]{zhu2024multi}
Zhu, H.; He, H.; Choromanska, A.; Ravindran, S.; Shi, B.; and Chen, L. 2024.
\newblock Multi-View Radar Autoencoder for Self-Supervised Automotive Radar Representation Learning.
\newblock In \emph{2024 IEEE Intelligent Vehicles Symposium (IV)}, 1601--1608. IEEE.

\end{thebibliography}


\newpage

\appendix
\twocolumn[ 
  \begin{center}
    \vspace*{1em}
    {\LARGE\bfseries Appendix}
    \vspace{1em}
  \end{center}
]

\newcommand{\cmark}{\ding{51}}   
\newcommand{\xmark}{\ding{55}}   

\setcounter{secnumdepth}{0} 

\section{Datasets Descriptions}
We use the KITTI3D, Waymo, and ONCE datasets to conduct self-supervised pre‐training and downstream fine‐tuning.

\textbf{KITTI3D} contains 3,712 training frames and 3,769 validation frames. We report the 3D object detection Average Precision (AP) metric at 40 recall positions ($R_{40}$) and 11 recall positions ($R_{11}$) for the moderate difficulty level\footnote{Moderate difficulty is defined in KITTI3D; LEVEL 2 difficulty, mentioned later, is defined in Waymo.} across all classes on the KITTI3D dataset.

\textbf{Waymo} contains 158,081 training frames and 39,987 validation frames. Following the standard OpenPCDet protocol, we subsample 20\% of the training frames at an interval of 5, yielding 31,617 frames. We report both the AP and the Average Precision with Heading (APH) metrics for LEVEL 2 difficulty across all classes on the Waymo dataset.

\textbf{ONCE} is a large‐scale unlabeled dataset designed for self‐supervised LiDAR pre‐training; it contains 1 million unlabeled frames. Depending on the subset, it offers small (100k), medium (500k), and large (1M) pools of unlabeled frames. In addition to this vast unlabeled corpus, it provides annotations for 16k LiDAR frames used in supervised downstream fine‐tuning.

\section{Experimental Details}
The random seed is set to 666 for all experiments.

\subsection{Pre-training Details}
We use our proposed AD-L-JEPA to pre-train the popular VoxelBackBone8x encoder, which contains $12$ sparse convolution layers. For all dataset settings, we pre-train the network with unlabeled training data for $30$ epochs using the Adam optimizer and the default one-cycle learning rate scheduler in OpenPCDet, with a learning rate of $0.0003$ and a weight decay of $0.01$. We set the initial $\eta$ to 0.996 and increase it linearly to $1$, for the moving average update. We set masking ratio = $0.5$. We set $\alpha_0=0.25$ and $\alpha_1=0.75$ for empty grids and non-empty, respectively. We set $\beta_1=1$ and $\beta_2=1$. The threshold for variance regularization, $\gamma$, is set to $\frac{1}{\sqrt{D}}=\frac{1}{16}$ for $L_2$-normalized embeddings. For the KITTI3D dataset, we set $\lambda_{\text{jepa}}=1$ and $\lambda_{\text{reg}}=1$ to balance the two loss functions. For all settings of the Waymo dataset and ONCE dataset, we set $\lambda_{\text{jepa}}=1$ and $\lambda_{\text{reg}}=10$ to balance the two loss functions. We use four 1080 Ti GPUs for KITTI3D with a total batch size of 16, and a single A100 GPU with a batch size of 16 for Waymo during pre-training. For the largest ONCE dataset 100k setting, we use 4 A100 GPUs with a total batch size of 64, we use 8 A100 GPUs with a total batch size of 128 for 500k and 1M settings.

For pre-training experiments reproducing Occupancy-MAE, we follow all default settings in their paper for both pre-training and fine-tuning. We use 4 1080 Ti GPUs for the KITTI3D dataset, 4 A100 GPUs for the Waymo dataset, and 4 A100 GPUs for the ONCE dataset.

\subsection{Fine-tuning Details}
During fine-tuning, we initialize the encoder with the pre-trained context encoder’s weights and fine-tune the model for 80 epochs using the SECOND, PVRCNN, and CenterPoint detectors (all without residual connections), employing the default OpenPCDet hyperparameters. For each setup, we conduct three independent runs and report the best performance. Reproduced results may differ from those in the original papers. We use four 1080 Ti GPUs for KITTI3D and four RTX 4090 GPUs for Waymo (batch size 16). For the largest ONCE experiments, we use eight RTX 4090 GPUs (total batch size 64, i.e., 4 per GPU).

\begin{figure*}[!ht]
    \centering
    \includegraphics[width=\linewidth]{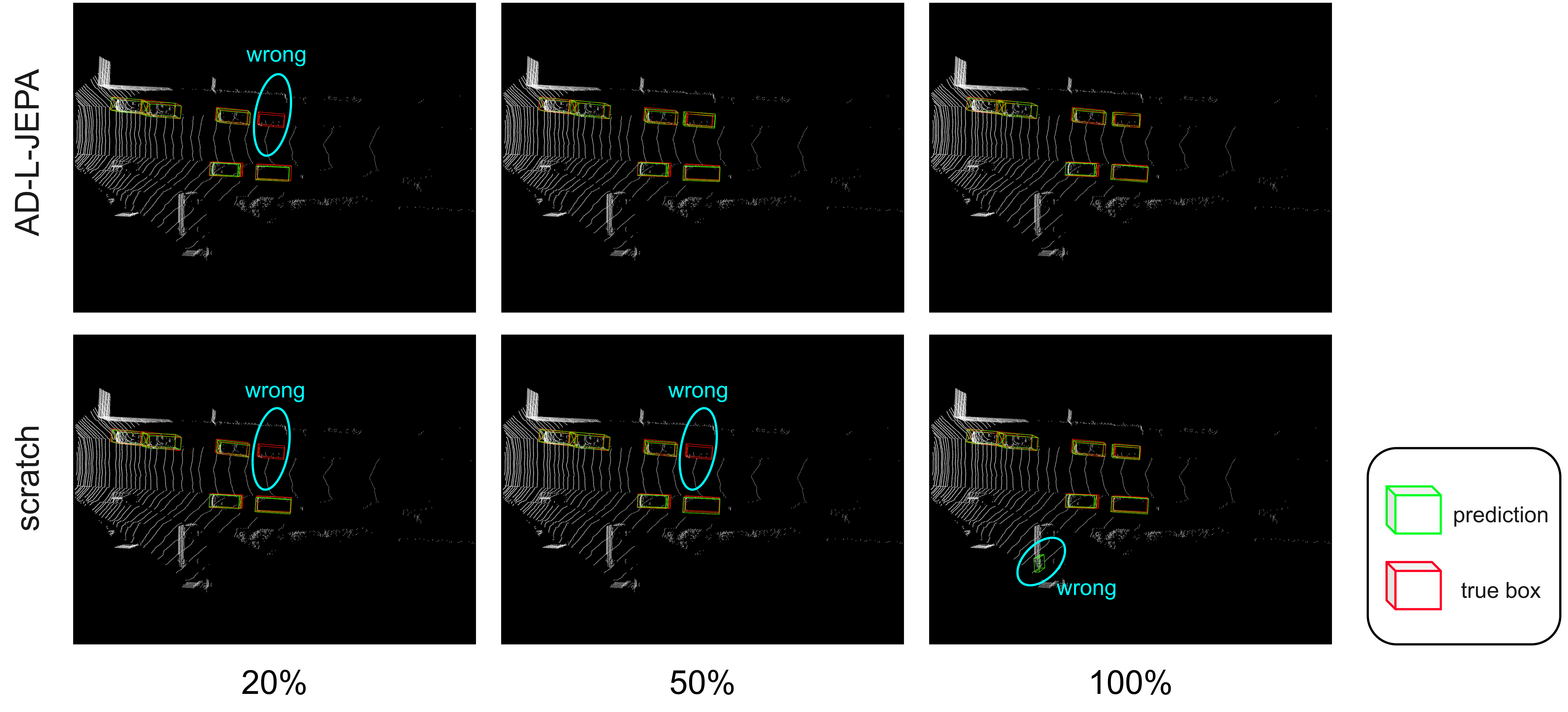}
    \caption{Visual comparison of 3D object detection performance between SECOND  trained from scratch and pre-trained with AD-L-JEPA, using 20\%, 50\%, and 100\% of labeled data from the KITTI3D data set.}
    \label{fig:visual_comparison}
\end{figure*}

\subsection{Transfer Learning and Label Effiency Details}
For transfer learning experiments from Waymo 20\% dataset to KITTI3D dataset: each point in the Waymo $20\%$ data set features five attributes: ['$x$', '$y$', '$z$', 'intensity', 'elongation'], whereas each point in the KITTI3D data set includes four attributes: ['$x$', '$y$', '$z$', 'intensity']. Consequently, the first layer of the encoder pre-trained on the Waymo data set has a shape of $[5, 16, 3, 3, 3]$, while the first layer used for object detection on the KITTI3D data set has a shape of $[4, 16, 3, 3, 3]$ to accommodate the respective point features with a sparse $3$D convolutional kernel of $3$. The remaining layers have the same shape in both the pre-trained and downstream networks. When fine-tuning the downstream network with an encoder pre-trained directly on the Waymo $20\%$ data set without modifications, the first layer is randomly initialized, while the remaining layers of the $3$D encoder retain the pre-trained weights.

Next, we also modify the first layer of the encoder used for pre-training on the Waymo $20\%$ data set to the shape of $[4, 16, 3, 3, 3]$ and redo the pre-training, utilizing only the first four features: `$x$', `$y$', `$z$', and `intensity', while ignoring the fifth feature `elongation'. This adjustment allows all layers of the encoder used for the downstream KITTI3D object detection task to be initialized from pre-trained weights. Furthermore, in this setting, for label efficiency of fine-tuning with varying amounts of labels, we report fine-tuning with $20\%$, $50\%$, and $100\%$ of the data and corresponding fine-tuning epochs of $320$, $160$, and $80$, respectively, to ensure a enough number of fine-tuning iterations.

\section{Additional Visualization}Figure~\ref{fig:visual_comparison} shows the qualitative comparisons with training from scratch using different amounts of labeled data for fine-tuning.

\section{Additional Ablation Studies}
In terms of computation, we conduct all ablation studies on the KITTI3D dataset for both pre-training and fine-tuning, using the simple SECOND model for downstream object detection.

\subsection{Ablation on Masking Ratios}
Table~\ref{table:ablation_masking} presents ablation studies with different masking ratios during pre-training and their corresponding impact on downstream object detection performance. We find that at a masking ratio of $0.5$, the difficulty is moderate and yields the best performance; thus, we set $0.5$ as the default masking ratio.
 
\begin{table}[ht]
    \centering
    \begin{tabular}{c|c|c|c}
        \hline
        masking ratio & 0.25 & 0.5 & 0.75\\
        \hline
        mAP - $R_{40}$  & 66.21 & \textbf{67.92} & 67.24\\
        \hline
        mAP - $R_{11}$ & 65.73 & \textbf{67.10} & 66.76\\
        \hline
    \end{tabular}
    \caption{Different masking ratio and the corresponding mAP (\%) on KITTI3D data set with SECOND detection model.}
    \label{table:ablation_masking}
\end{table}

\subsection{Ablation on Learnable Empty/Mask Tokens and Explict Variance Regularization Loss}

Please see the ablation studies in Table~\ref{table:ablations}. Replacing representations of masked regions with a learnable mask token is a common practice in self-supervised learning. In LiDAR settings using a sparse convolutional network, empty regions (i.e., regions with no points) produce near-zero vectors after the target encoder, resulting in semantically similar embeddings. Without replacing these with a learnable empty token, direct $L_2$ normalization of such near-zero vectors yields high variance in the embeddings and causes them to diverge from their intended meanings. Additionally, when using explicit variance loss to prevent representation collapse, the downstream fine-tuning performance significantly improves. Given the above ablation studies, we introduce a learnable empty token, a learnable mask token, and explicit variance regularization as the default setting.

\begin{table}[ht!]
    \centering
    \small
    \begin{tabular}{c|c|c|c|c}
        \toprule
        Learnable empty token & \cmark & \xmark & \cmark  & \cmark\\
        Learnable mask token  & \cmark & \cmark & \xmark  & \cmark\\
        Variance regularization loss & \xmark & \cmark & \cmark  & \cmark\\
        \midrule
        mAP - $R_{40}$             &     65.86 & 66.37    &   67.20  & \textbf{67.92} \\
        mAP - $R_{11}$    & 65.48         &     65.91   &   66.77    & \textbf{67.10} \\
        \bottomrule
    \end{tabular}
    \vspace{-0.1in}
    \caption{Ablation study of variance regularization loss, learnable empty token, and learnable mask token during pre-training, along with the corresponding fine-tuning mAP (\%) on the KITTI3D dataset using the SECOND detection model.}
    \label{table:ablations}
\end{table}

\subsection{Ablation on Predictor's Depth}
As the encoder's architecture is fixed and thus can't be changed for all settings, we conduct ablation studies on varying the predictor's depth in Table~\ref{tab:ablation_predictor_depth} and observe empirically that depth=3 works the best. Therefore, we use that as the default setting for all experiments.

\begin{table}[ht!]
    \centering
    \small
    \begin{tabular}{c|c|c|c}
        \toprule
        Number of layers & 1 & 3 & 6  \\
        \midrule
        mAP - $R_{40}$             &    67.58   &  \textbf{67.92}   &    67.17  \\
        \hline
        mAP - $R_{11}$    &    66.94    &  \textbf{67.10}      &    66.45  \\
        \bottomrule
    \end{tabular}
    \vspace{-0.1in}
    \caption{Ablation study of predictor depth during pre-training, along with the corresponding fine-tuning mAP (\%) on the KITTI3D dataset using the SECOND detection model.}
    \label{tab:ablation_predictor_depth}
\end{table}

\section{Detailed Pre-training Efficiency Comparisons}

The detailed comparison of per-GPU memory consumption versus batch size between Occupancy-MAE and AD-L-JEPA can be found in Tables~\ref{tab:gpu_waymo} and~\ref{tab:gpu_once}. AD-L-JEPA leads to $2.8\times$–$3.4\times$ lower GPU memory usage for pre‑training on the 20\% and 100\% splits of the Waymo dataset, and $3.1\times$–$4\times$ lower GPU memory usage for pre‑training on the ONCE 100k split.

\subsection{Pre-training GPU Memory Comparison}

\begin{table}[ht!]
    \centering
    \small
    \begin{tabular}{c|c|c|c}
        \toprule
        Batch size & 4 & 8 & 16  \\
        \midrule
        Occupancy-MAE             &    18.4 GB &   34.6 GB &   67.1 GB \\
        \hline
        AD-L-JEPA (ours)     &   \textbf{6.7  GB}  &  \textbf{10.7  GB}   &    \textbf{19.6 GB} \\
        \bottomrule
    \end{tabular}
    \vspace{-0.1in}
    \caption{Per-GPU memory (GB) consumption as batch size increases during pre-training on the Waymo dataset, measured on an A100 GPU.}
    \label{tab:gpu_waymo}
\end{table}

\begin{table}[ht!]
    \centering
    \small
    \begin{tabular}{c|c|c|c}
        \toprule
        Batch size & 4 & 8 & 16  \\
        \midrule
        Occupancy-MAE             &     17.7 GB &  33.7 GB &  65.4 GB \\
        \hline
        AD-L-JEPA (ours)     &   \textbf{5.7  GB}  &  \textbf{  9 GB}   &    \textbf{16.5 GB} \\
        \bottomrule
    \end{tabular}
    \vspace{-0.1in}
    \caption{Per-GPU memory (GB) consumption as batch size increases during pre-training on the ONCE dataset, measured on an A100 GPU.}
    \label{tab:gpu_once}
\end{table}

\subsection{Pre-training Time Comparison}
Although, as mentioned in the previous section, AD-L-JEPA is efficient enough so that it can use only one A100 GPU for pre-training on the medium-scale Waymo dataset (whereas Occupancy-MAE requires four), we estimate the pre-training hours for both methods using 4 A100 GPUs on the Waymo dataset to enable a fair comparison. For the ONCE 100k dataset, we estimate the pre‐training hours for both methods using eight A100 GPUs. The detailed comparison of pre-training A100 hours consumption can be found in Tables~\ref{tab:gpu_hour}. AD-L-JEPA reduces pre-training GPU hours by $2.7\times$ on the 20\% and 100\% Waymo splits and by $1.9\times$ on the ONCE 100k split.

\begin{table}[ht!]
    \centering
    \small
    \begin{tabular}{c|c|c|c}
        \toprule
        Dataset & Waymo 20\% & Waymo 100\% & ONCE 100k \\
        \midrule
        Occupancy-MAE             &    56 hrs & 210 hrs & 56 hrs \\
        \hline
        AD-L-JEPA (ours)     &   \textbf{19.25 hrs}  &  \textbf{77 hrs}   &    \textbf{30 hrs} \\
        \bottomrule
    \end{tabular}
    \vspace{-0.1in}
    \caption{Pre‐training A100 GPU hours for the Waymo 20\%, Waymo 100\%, and ONCE 100k datasets.}
    \label{tab:gpu_hour}
\end{table}

\section{Pre-training logs}
\label{sec:pretraining_logs}

In this section, we present the pre-training logs over the course of training iterations for each setting. 

\begin{figure*}[ht!]
    \centering
    \includegraphics[width=0.9\linewidth]{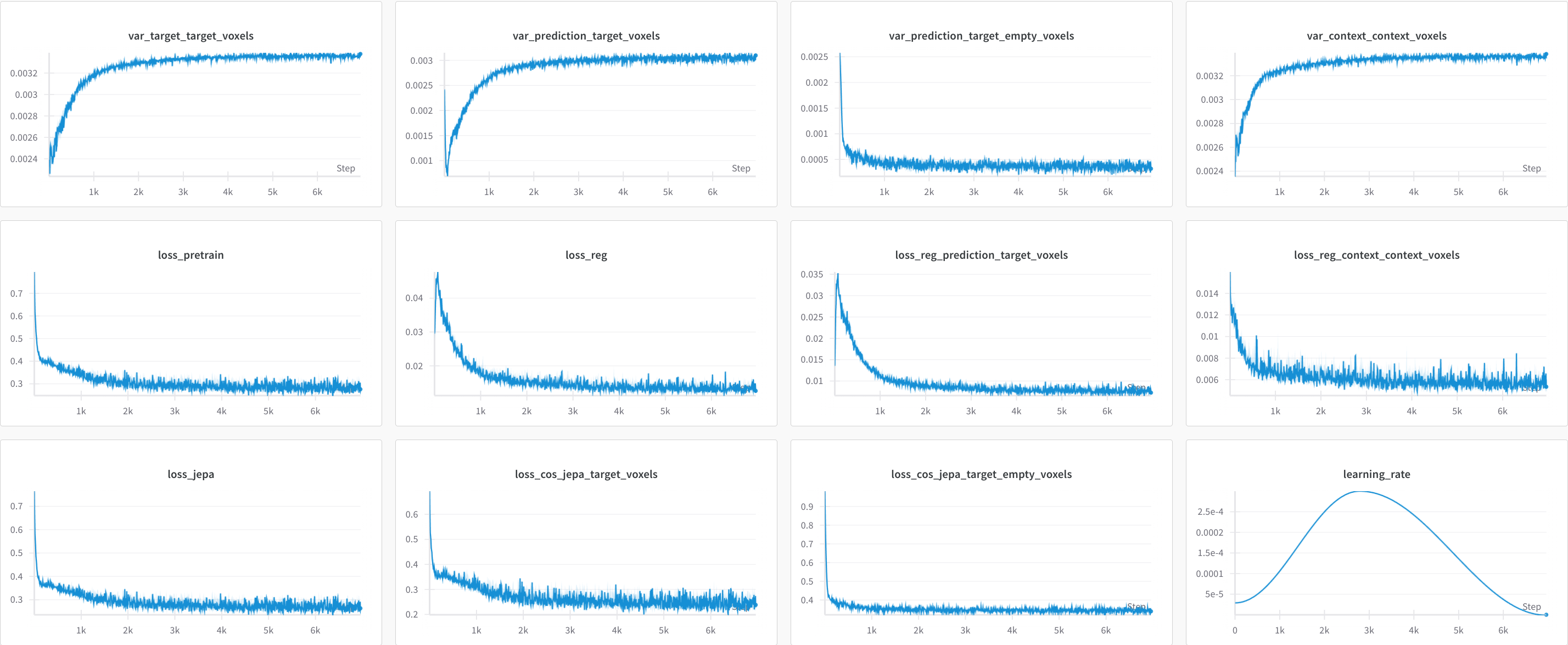}
    \caption{Pretraining logs using AD-L-JEPA on the KITTI3D dataset (best viewed when zoomed in).}
    \label{fig:log_kitti}
\end{figure*}

\begin{figure*}[ht!]
    \centering
    \includegraphics[width=0.9\linewidth]{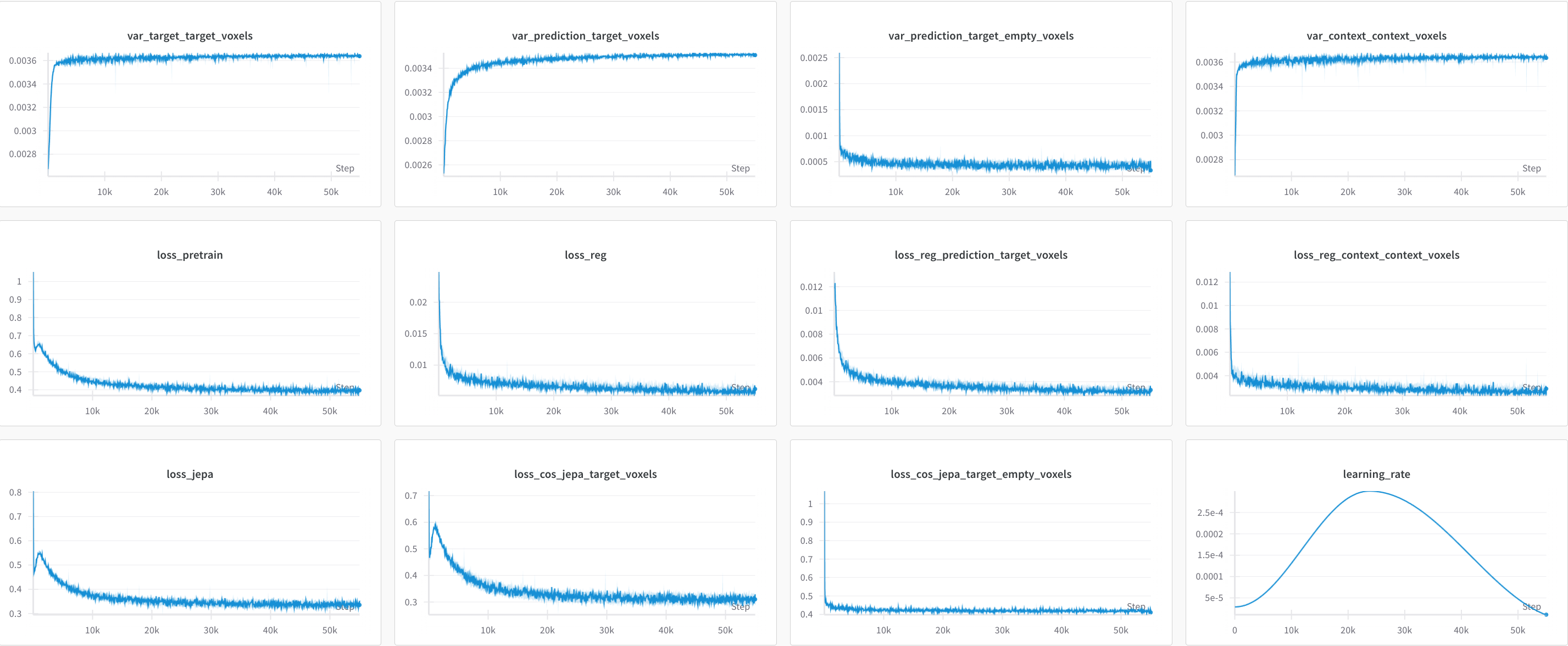}
    \caption{Pretraining logs using AD-L-JEPA on the Waymo 20\% dataset (best viewed when zoomed in).}
    \label{fig:log_waymo_20}
\end{figure*}

\begin{figure*}[ht!]
    \centering
    \includegraphics[width=0.9\linewidth]{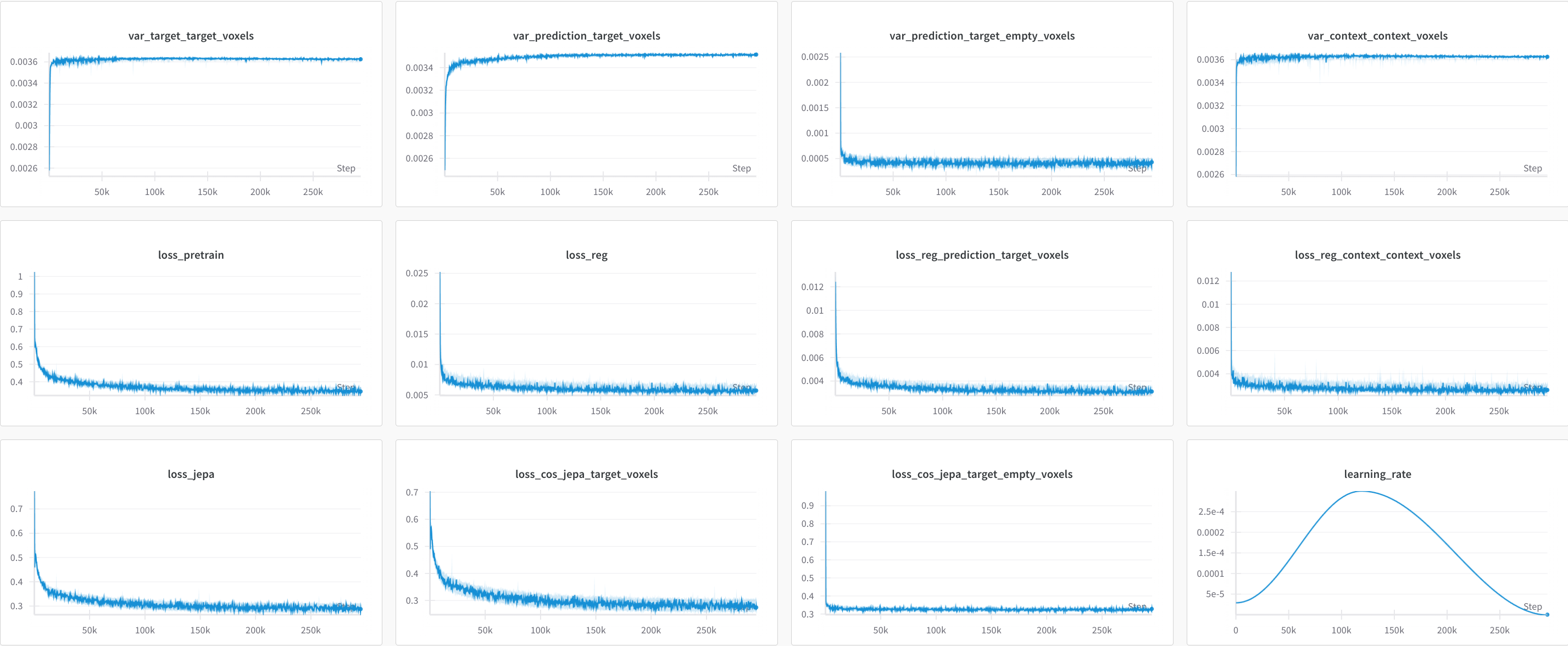}
    \caption{Pretraining logs using AD-L-JEPA on the Waymo 100\% dataset (best viewed when zoomed in).}
    \label{fig:log_waymo_100}
    \vspace{-0.1in}
\end{figure*}

\begin{figure*}[ht!]
    \centering
    \includegraphics[width=0.8\linewidth]{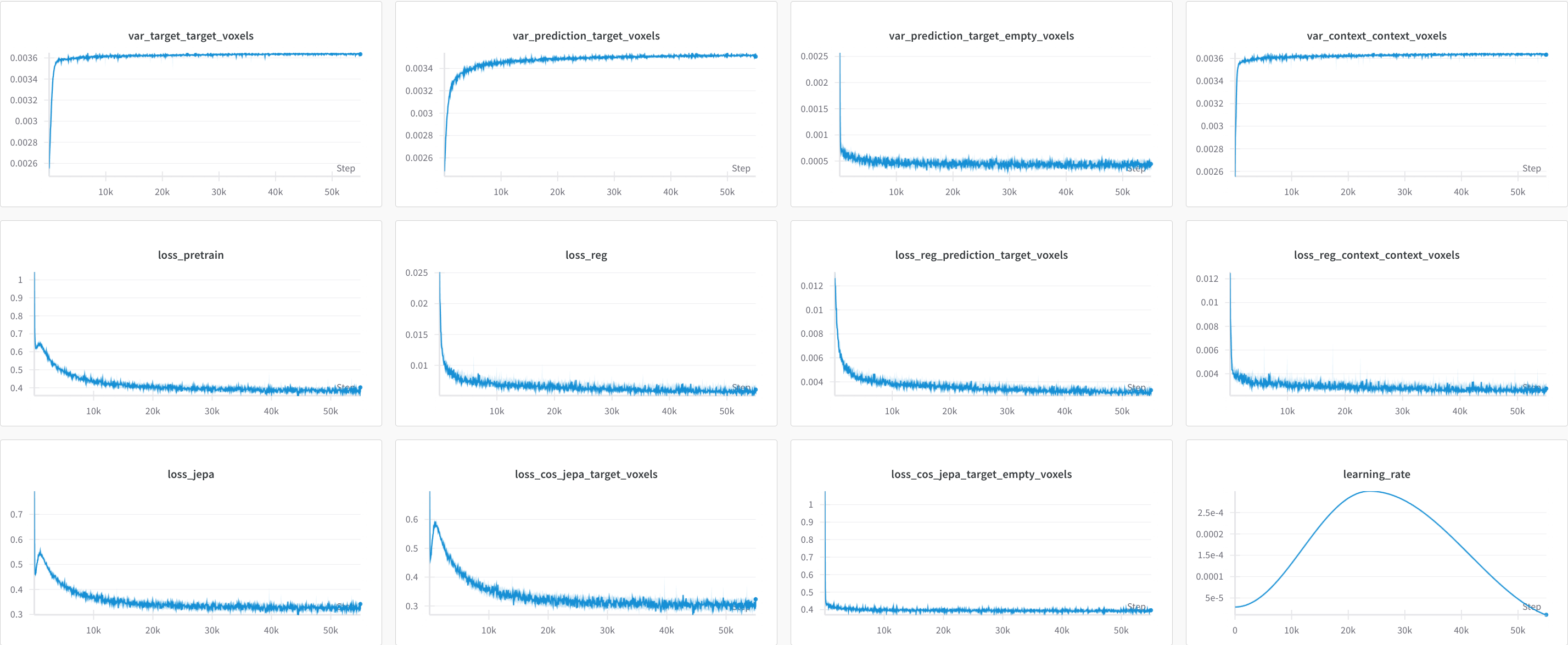}
    \caption{Pretraining logs using AD-L-JEPA on the Waymo 20\% dataset for transfer to downstream tasks on the KITTI3D dataset (best viewed when zoomed in). Only the first four point features---\(\mathit{x}\), \(\mathit{y}\), \(\mathit{z}\), and intensity---are used, ignoring the fifth feature, elongation.}
    \label{fig:log_waymo_20_transfer}
\end{figure*}

For pre-training experiments on the KITTI3D dataset, each point has four features ['$x$', '$y$', '$z$', 'intensity'], and the corresponding pre-training logs are shown in Figure~\ref{fig:log_kitti}; For pre-training experiments on the Waymo 20\% and 100\% datasets, each point has five features ['$x$', '$y$', '$z$', 'intensity', 'elongation']. The logs for these experiments appear in Figure~\ref{fig:log_waymo_20} and Figure~\ref{fig:log_waymo_100}, respectively.; Additionally, we conduct an experiment that uses only the first four point features ['$x$', '$y$', '$z$', 'intensity'], excluding the fifth feature (elongation). Under this setting, the first layer of the pretrained encoder can be directly reused together with the other pretrained layers during fine-tuning. The logs for this experiment are shown in Figure~\ref{fig:log_waymo_20_transfer}.

\noindent
In each setting, we track 12 values over the training iterations: 
\begin{itemize}
    \item \texttt{loss\_pretrain}: overall pre-training loss
    \item \texttt{loss\_reg}: overall variance regularization loss
    \item \texttt{loss\_reg\_prediction\_target\_voxels}: variance regularization loss for predicted target BEV embeddings at non-empty regions
    \item \texttt{loss\_reg\_context\_context\_voxels}: variance regularization loss for context BEV embeddings at non-empty regions
    \item \texttt{loss\_jepa}: overall joint-embedding prediction loss
    \item \texttt{loss\_cos\_jepa\_target\_voxels}: joint-embedding prediction loss for non-empty regions
    \item \texttt{loss\_cos\_jepa\_target\_empty\_voxels}: joint-embedding prediction loss for empty regions
    \item \texttt{var\_target\_target\_voxels}: variance for non-empty regions in target BEV embeddings
    \item \texttt{var\_prediction\_target\_voxels}: variance for non-empty regions in predicted target BEV embeddings 
    \item \texttt{var\_prediction\_target\_empty\_voxels}: variance for empty regions in predicted target BEV embeddings
    \item \texttt{var\_context\_context\_voxels}: variance for non-empty regions in context BEV embeddings
    \item \texttt{learning\_rate}: learning rate
\end{itemize}

\begin{figure*}[ht!]
    \centering
    \includegraphics[width=0.8\linewidth]{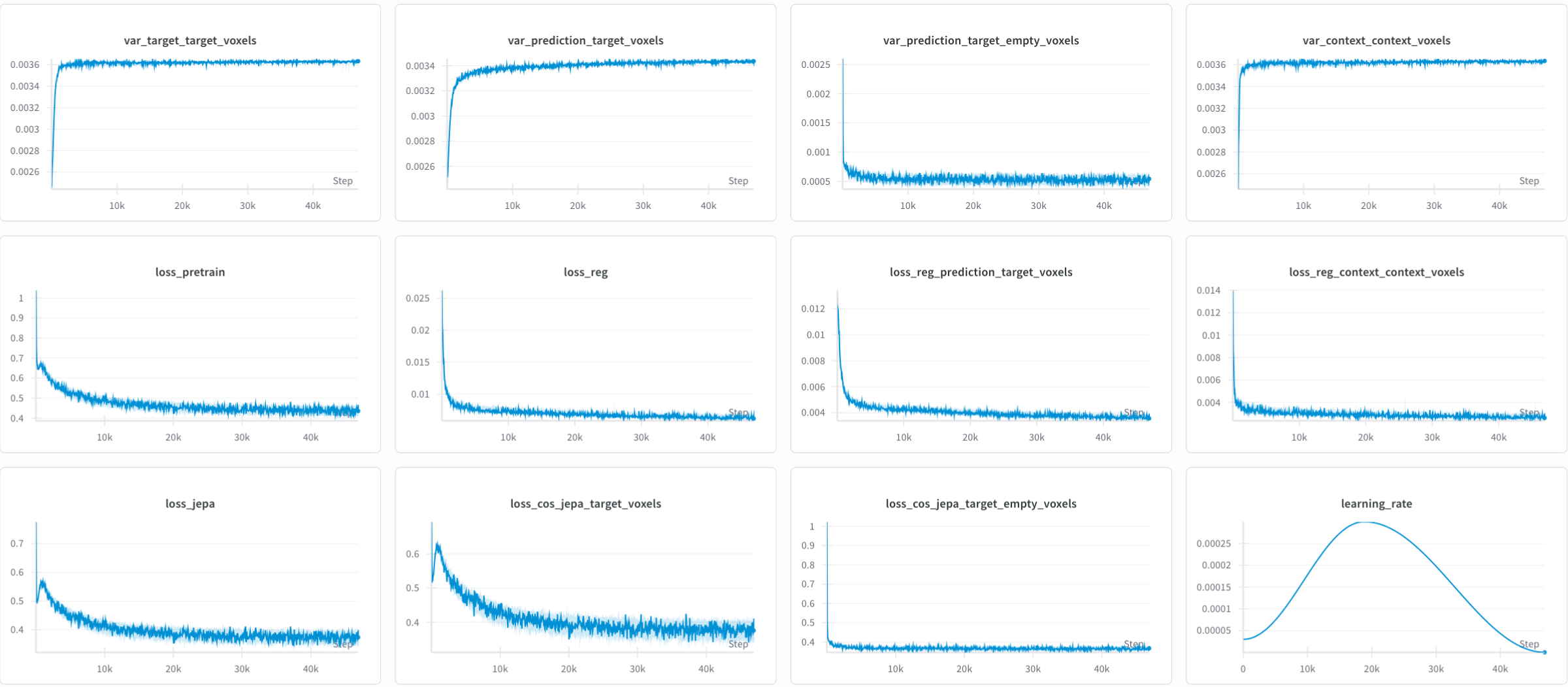}
    \caption{Pretraining logs using AD-L-JEPA on the ONCE 100k dataset (best viewed when zoomed in).}
    \label{fig:log_once_100k}
\end{figure*}

\begin{figure*}[ht!]
    \centering
    \includegraphics[width=0.8\linewidth]{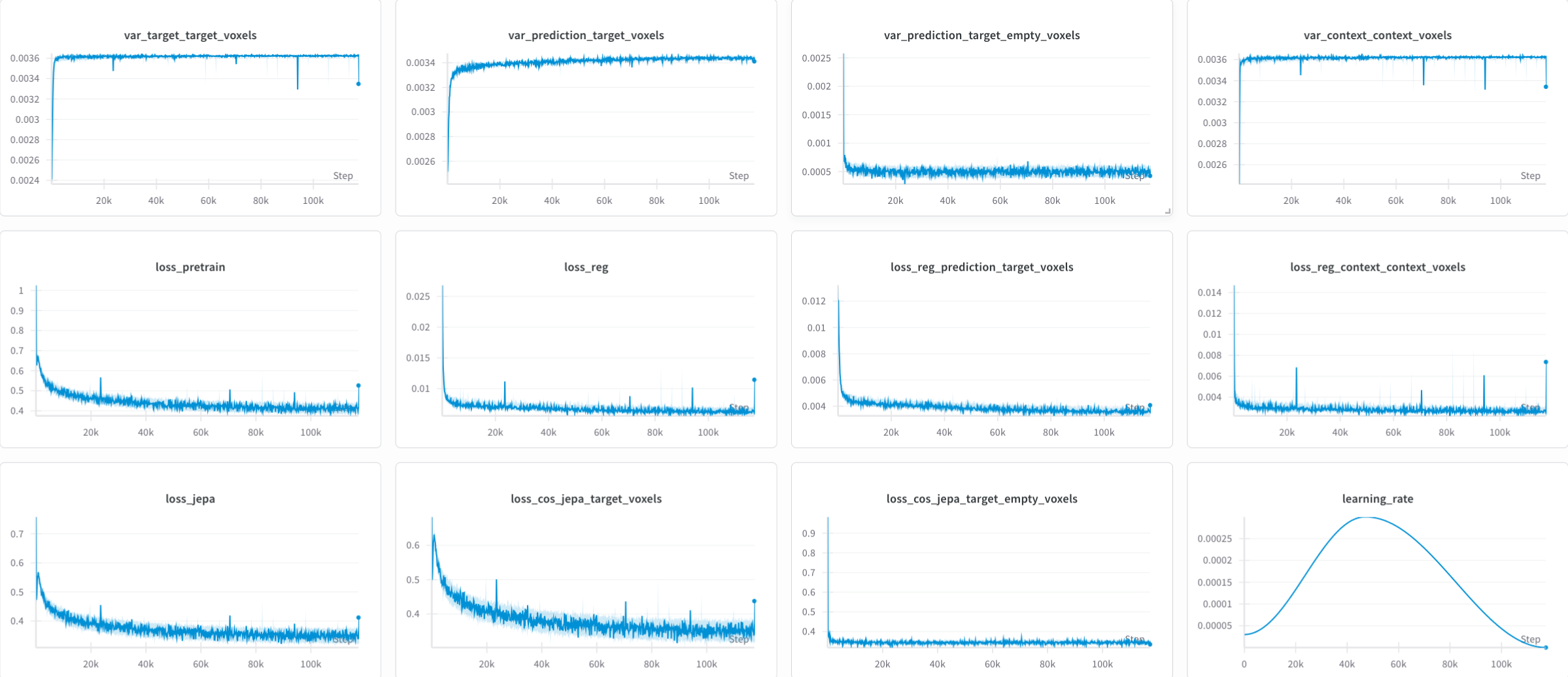}
    \caption{Pretraining logs using AD-L-JEPA on the ONCE 500k dataset (best viewed when zoomed in).}
    \label{fig:log_once_500k}
\end{figure*}

\begin{figure*}[ht!]
    \centering
    \includegraphics[width=0.9\linewidth]{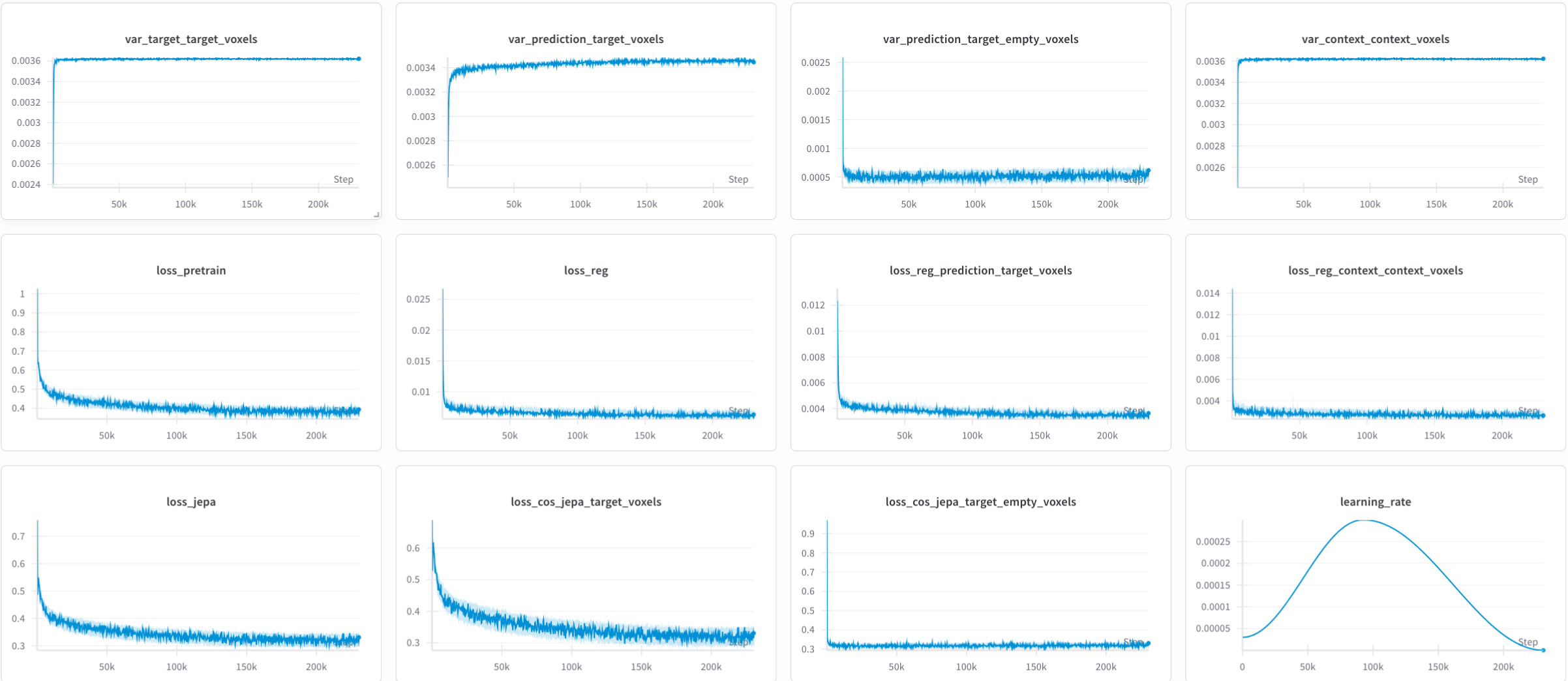}
    \caption{Pretraining logs using AD-L-JEPA on the ONCE 1M dataset (best viewed when zoomed in).}
    \label{fig:log_once_1M}
\end{figure*}

\section{Detailed Downstream Fine-tuning Results}
\begin{table*}[ht!]
\small
\centering
\setlength{\tabcolsep}{4pt}
\resizebox{\textwidth}{!}{
\begin{tabular}{lll|ccc|ccc|ccc}
\toprule
Backbone & Metric & Method & \multicolumn{3}{c|}{Car} & \multicolumn{3}{c|}{Pedestrian} & \multicolumn{3}{c}{Cyclist} \\
& & & Easy & Mod. & Hard &Easy& Mod. & Hard & Easy& Mod. & Hard\\
\midrule
$R_{40}$ metric & 2D object detection 
& No pre-training     & {96.29} & 94.05 & 91.63 & {68.61} & {65.73} & {62.17} & 89.06 & 77.96 & 73.99 \\
\cmidrule{3-12}
&& ALSO             & 95.75 & {94.45} & \textbf{91.89} & 67.63 & 64.75 & 61.66 & 91.99 & \textbf{81.87} & \textbf{78.42} \\
&& Occupancy-MAE    & 95.80 & \textbf{94.52} & {91.88} & 67.16 & 64.61 & 62.07 & {93.01} & 80.04 & 76.11 \\
&& AD-L-JEPA        & \textbf{96.90} & 94.31 & 91.85 & \textbf{72.14} & \textbf{68.62} & \textbf{65.49} & \textbf{93.26} & {81.06} & {77.25} \\

\cmidrule{2-12}
&Bird's eye view 
& No pre-training     & \textbf{93.27} & \textbf{90.28} & \textbf{87.85} & {63.37} & {58.30} & {53.94} & 85.58 & 70.49 & 66.03 \\
\cmidrule{3-12}
&& ALSO             & 92.28 & 89.46 & 87.44 & 61.72 & 57.06 & 52.70 & 86.48 & 72.04 & 67.63 \\
&& Occupancy-MAE    & {92.42} & 89.64 & 87.33 & 59.74 & 55.91 & 51.80 & \textbf{89.16} & \textbf{73.17} & \textbf{68.71} \\
&& AD-L-JEPA        & 92.15 & {90.10} & {87.61} & \textbf{64.20} & \textbf{59.67} & \textbf{55.06} & {88.74} & {72.80} & {68.42} \\

\cmidrule{2-12}
&3D object detection
& No pre-training     & {90.46} & \textbf{81.99} & \textbf{79.05} & {57.12} & 52.02 & 47.53 & 80.36 & 65.07 & 60.99 \\
\cmidrule{3-12}
&& ALSO             & 89.91 & 81.48 & 78.50 & 56.96 & {52.50} & {47.93} & 81.51 & 65.95 & 61.88 \\
&& Occupancy-MAE    & 88.59 & 81.15 & 78.04 & 54.68 & 50.36 & 46.16 & \textbf{86.55} & \textbf{69.74} & \textbf{65.50} \\
&& AD-L-JEPA        & \textbf{90.47} & {81.68} & {78.59} & \textbf{58.76} & \textbf{54.15} & \textbf{49.39} & {85.33} & {67.93} & {64.07} \\

\cmidrule{2-12}
&Orientation similarity
& No pre-training     & {96.28} & 93.92 & 91.42 & {63.41} & {60.21} & {56.35} & 88.68 & 76.90 & 72.93 \\
\cmidrule{3-12}
&& ALSO             & 95.73 & {94.28} & 91.61 & 62.60 & 59.22 & 55.79 & 91.81 & \textbf{80.66} & \textbf{77.08} \\
&& Occupancy-MAE    & 95.78 & \textbf{94.37} & {91.62} & 62.02 & 58.77 & 55.91 & {92.72} & 79.37 & 75.38 \\
&& AD-L-JEPA        & \textbf{96.88} & 94.17 & \textbf{91.64} & \textbf{67.79} & \textbf{63.79} & \textbf{60.29} & \textbf{93.13} & {80.33} & {76.38} \\

\midrule\midrule
$R_{11}$ metric & 2D object detection
& No pre-training     & {94.10} & 89.88 & 89.03 & {68.18} & {65.61} & 62.15 & 85.94 & 76.27 & 72.67 \\
\cmidrule{3-12}
&& ALSO             & 90.82 & {90.04} & \textbf{89.21} & 67.30 & 64.69 & 62.08 & 89.58 & {80.10} & \textbf{77.46} \\
&& Occupancy-MAE    & 90.83 & \textbf{90.06} & {89.17} & 66.10 & 64.26 & {62.39} & {90.02} & 78.09 & 74.53 \\
&& AD-L-JEPA        & \textbf{94.79} & 89.98 & 89.17 & \textbf{70.88} & \textbf{67.87} & \textbf{65.69} & \textbf{91.64} & \textbf{80.14} & {75.91} \\

\cmidrule{2-12}
&Bird's eye view
& No pre-training     & \textbf{90.02} & \textbf{88.20} & \textbf{86.81} & {63.78} & {58.73} & {55.46} & 83.26 & 69.84 & 65.82 \\
\cmidrule{3-12}
&& ALSO             & {89.93} & 87.78 & 86.21 & 62.35 & 57.46 & 53.93 & 83.11 & 70.63 & {67.46} \\
&& Occupancy-MAE    & 89.87 & 87.75 & 86.17 & 60.62 & 56.60 & 53.04 & \textbf{86.74} & {71.58} & \textbf{68.09} \\
&& AD-L-JEPA        & 89.70 & {87.92} & {86.48} & \textbf{64.05} & \textbf{59.32} & \textbf{55.76} & {84.94} & \textbf{72.04} & 67.45 \\

\cmidrule{2-12}
& 3D object detection
& No pre-training     & \textbf{88.47} & \textbf{78.89} & \textbf{77.69} & 57.33 & 53.78 & 48.79 & 79.79 & 64.93 & 60.55 \\
\cmidrule{3-12}
&& ALSO             & {88.34} & 78.37 & 77.13 & {57.39} & {53.94} & {49.19} & 80.17 & 66.06 & 61.36 \\
&& Occupancy-MAE    & 87.71 & 78.16 & 76.69 & 55.12 & 52.10 & 47.95 & \textbf{84.46} & \textbf{69.39} & \textbf{65.31} \\
&& AD-L-JEPA        & 88.27 & {78.51} & {77.19} & \textbf{58.35} & \textbf{54.86} & \textbf{50.40} & {82.07} & {67.94} & {63.91} \\

\cmidrule{2-12}
& Orientation similarity
& No pre-training     & {94.08} & 89.77 & 88.85 & {63.63} & {60.70} & {56.97} & 85.59 & 75.27 & 71.73 \\
\cmidrule{3-12}
&& ALSO             & 90.81 & {89.90} & {88.96} & 62.64 & 59.69 & 56.67 & 89.42 & {79.03} & \textbf{76.28} \\
&& Occupancy-MAE    & 90.82 & \textbf{89.94} & 88.94 & 61.29 & 58.81 & 56.48 & {89.75} & 77.47 & 73.85 \\
&& AD-L-JEPA        & \textbf{94.77} & 89.87 & \textbf{88.98} & \textbf{66.95} & \textbf{63.55} & \textbf{60.99} & \textbf{91.50} & \textbf{79.42} & {75.12} \\
\bottomrule
\end{tabular}
}

\caption{Detection results using SECOND on the KITTI3D validation set for all difficulty levels, 
reported with \(R_{40}\) and \(R_{11}\) AP~(\%) across all metrics. The models are pre-trained 
and then fine-tuned on the KITTI3D dataset. Following standard practice, we report the 
'3D object detection' metric in the main paper.
}
\label{tab:adetails_kitti3d_second}
\end{table*}

\begin{table*}[ht!]
\small
\centering
\setlength{\tabcolsep}{4pt}

\begin{tabular}{lll|ccc|ccc|ccc}
\toprule
Backbone & Metric & Method & \multicolumn{3}{c|}{Car} & \multicolumn{3}{c|}{Pedestrian} & \multicolumn{3}{c}{Cyclist} \\
& & & Easy & Mod. & Hard &Easy& Mod. & Hard & Easy& Mod. & Hard\\
\midrule
$R_{40}$ metric & 2D object detection 
& No pre-training     & 97.50 & 94.45 & {94.06} & {74.14} & {68.84} & 65.91 & 94.66 & {82.16} & {79.35} \\
\cmidrule{3-12}
&& ALSO             & {98.18} & \textbf{94.55} & \textbf{94.14} & 72.97 & 68.40 & {65.99} & \textbf{97.48} & \textbf{83.20} & \textbf{80.51} \\
&& Occupancy-MAE    & \textbf{98.23} & {94.47} & 93.93 & 73.66 & 68.67 & 65.35 & {96.00} & 80.53 & 77.54 \\
&& AD-L-JEPA        & 98.18 & 94.43 & 94.03 & \textbf{76.59} & \textbf{70.43} & \textbf{67.32} & 95.62 & 82.08 & 78.81 \\

\cmidrule{2-12}
&Bird's eye view 
& No pre-training     & 93.03 & 90.56 & 88.51 & 66.84 & 59.22 & 54.89 & 93.88 & 74.69 & {71.41} \\
\cmidrule{3-12}
&& ALSO             & 93.00 & {90.60} & {88.63} & 66.50 & 59.79 & 55.65 & \textbf{94.74} & \textbf{76.39} & \textbf{72.01} \\
&& Occupancy-MAE    & \textbf{94.55} & 90.53 & 88.51 & {68.64} & {61.00} & {56.35} & 92.05 & {74.82} & 70.39 \\
&& AD-L-JEPA        & {94.47} & \textbf{90.79} & \textbf{88.64} & \textbf{70.38} & \textbf{62.86} & \textbf{58.05} & {93.89} & 74.77 & 71.11 \\

\cmidrule{2-12}
&3D object detection
& No pre-training     & 92.02 & {84.65} & 82.49 & 63.70 & 56.19 & 51.55 & 91.48 & 72.19 & {69.15} \\
\cmidrule{3-12}
&& ALSO             & {92.13} & 84.64 & {82.53} & 64.26 & 57.09 & {52.87} & \textbf{92.33} & \textbf{73.72} & \textbf{69.32} \\
&& Occupancy-MAE    & 91.57 & 84.34 & 82.29 & {65.36} & {57.55} & 52.02 & 90.66 & 71.33 & 66.95 \\
&& AD-L-JEPA        & \textbf{92.40} & \textbf{85.07} & \textbf{82.78} & \textbf{67.19} & \textbf{59.68} & \textbf{54.08} & {91.61} & {73.02} & 68.50 \\

\cmidrule{2-12}
&Orientation similarity
& No pre-training     & 97.48 & 94.31 & {93.84} & {68.82} & 63.20 & 59.91 & 94.47 & {81.59} & {78.56} \\
\cmidrule{3-12}
&& ALSO             & 98.14 & \textbf{94.41} & \textbf{93.90} & 68.22 & {63.57} & {60.76} & \textbf{97.22} & \textbf{82.65} & \textbf{79.80} \\
&& Occupancy-MAE    & \textbf{98.20} & {94.34} & 93.70 & 68.60 & 63.16 & 59.67 & {95.82} & 80.20 & 77.12 \\
&& AD-L-JEPA        & {98.15} & 94.30 & 93.80 & \textbf{71.06} & \textbf{65.08} & \textbf{61.70} & 95.32 & 81.24 & 77.88 \\

\midrule\midrule
$R_{11}$ metric & 2D object detection
& No pre-training     & \textbf{95.79} & 89.56 & 89.21 & {73.49} & 68.50 & 65.17 & 91.68 & 80.51 & {76.59} \\
\cmidrule{3-12}
&& ALSO             & 95.59 & \textbf{89.69} & \textbf{89.26} & 72.44 & 68.03 & 65.29 & \textbf{96.42} & \textbf{81.87} & \textbf{77.54} \\
&& Occupancy-MAE    & {95.67} & 89.57 & {89.23} & 73.02 & \textbf{68.95} & {65.30} & {92.95} & 77.60 & 75.97 \\
&& AD-L-JEPA        & 95.52 & {89.58} & 89.17 & \textbf{75.19} & {68.92} & \textbf{67.00} & 92.92 & {80.82} & 76.53 \\

\cmidrule{2-12}
&Bird's eye view
& No pre-training     & 90.22 & 88.03 & 87.48 & 65.97 & 60.03 & 55.25 & {93.34} & {73.83} & {70.61} \\
\cmidrule{3-12}
&& ALSO             & {90.23} & {88.13} & {87.56} & 65.59 & 60.79 & 55.84 & \textbf{93.77} & \textbf{74.06} & \textbf{71.35} \\
&& Occupancy-MAE    & 90.02 & 88.03 & 87.47 & \textbf{68.82} & {61.86} & {56.45} & 87.74 & 73.65 & 70.06 \\
&& AD-L-JEPA        & \textbf{90.30} & \textbf{88.20} & \textbf{87.60} & {68.66} & \textbf{62.91} & \textbf{57.61} & 91.18 & 73.81 & 70.11 \\

\cmidrule{2-12}
& 3D object detection
& No pre-training     & 89.43 & 83.38 & 78.79 & 63.92 & 57.01 & 52.81 & 86.63 & {72.65} & {68.64} \\
\cmidrule{3-12}
&& ALSO             & {89.48} & {83.48} & {78.86} & 64.43 & 57.99 & {54.03} & {87.30} & \textbf{72.89} & \textbf{69.43} \\
&& Occupancy-MAE    & 88.94 & 83.41 & 78.67 & {65.05} & {58.68} & 53.72 & 86.13 & 70.97 & 65.28 \\
&& AD-L-JEPA        & \textbf{89.72} & \textbf{83.74} & \textbf{79.01} & \textbf{66.64} & \textbf{59.91} & \textbf{55.45} & \textbf{87.32} & 72.19 & 68.49 \\

\cmidrule{2-12}
& Orientation similarity
& No pre-training     & \textbf{95.76} & 89.47 & 89.04 & {68.65} & 63.23 & 59.78 & 91.48 & {80.01} & {75.93} \\
\cmidrule{3-12}
&& ALSO             & 95.55 & \textbf{89.58} & \textbf{89.07} & 67.98 & 63.49 & {60.65} & \textbf{96.16} & \textbf{81.32} & \textbf{76.95} \\
&& Occupancy-MAE    & {95.64} & 89.47 & {89.06} & 68.37 & {64.00} & 60.18 & {92.75} & 77.33 & 75.59 \\
&& AD-L-JEPA        & 95.48 & {89.48} & 89.01 & \textbf{70.30} & \textbf{64.31} & \textbf{61.99} & 92.66 & 80.01 & 75.74 \\
\bottomrule
\end{tabular}

\caption{Detection results using PV-RCNN on the KITTI3D validation set for all difficulty levels, 
reported with \(R_{40}\) and \(R_{11}\) AP~(\%) across all metrics. The models are pre-trained 
and then fine-tuned on the KITTI3D dataset. Following standard practice, we report the 
'3D object detection' metric in the main paper.}
\label{tab:adetails_kitti3d_pv_rcnn}
\end{table*}

\begin{table*}[ht!]
\small
\centering
\setlength{\tabcolsep}{4pt}
\resizebox{\textwidth}{!}{
\begin{tabular}{l|cccccccccccc}
\toprule
Method & \multicolumn{4}{c}{Vehicle} & \multicolumn{4}{c}{Pedestrian} & \multicolumn{4}{c}{Cyclist} \\
 & L1 AP & L1 APH & L2 AP & L2 APH & L1 AP & L1 APH & L2 AP & L2 APH & L1 AP & L1 APH & L2 AP & L2 APH \\
\midrule
No pre-training & 71.37 & 70.80 & 63.28 & 62.77 & 71.86 & 65.32 & 63.95 & 57.98 & 69.34 & 68.06 & 66.78 & 65.55 \\
Occupancy-MAE, 20\% & 71.27 & 70.72 & 63.20 & 62.70 & 72.20 & 65.72 & 64.20 & 58.29 & 69.77 & 68.52 & 67.20 & 66.00 \\
Occupancy-MAE, 100\% & {71.59} & {71.04} & {63.53} & {63.04} & \textbf{72.63} & \textbf{66.16} & \textbf{64.73} & \textbf{58.81} & {70.38} & {69.12} & {67.77} & {66.55} \\
AD-L-JEPA (ours), 20\% & 71.24 & 70.69 & 63.18 & 62.68 & 72.30 & 65.77 & 64.35 & 58.39 & 70.26 & 69.01 & 67.68 & 66.48 \\
AD-L-JEPA (ours), 100\% & \textbf{71.67} & \textbf{71.12} & \textbf{63.58} & \textbf{63.07} & {72.57} & {66.06} & {64.58} & {58.64} & \textbf{70.69} & \textbf{69.38} & \textbf{68.07} & \textbf{66.81} \\
\bottomrule
\end{tabular}
}

\caption{Detection results using Centerpoint on the Waymo validation set for all difficulty levels, 
reported with AP~(\%) and  APH~(\%) metrics. The models are pre-trained with $20\%$ or $100\%$ of the Waymo data set, as indicated, and are fine-tuned on the Waymo 20\% dataset. Following standard practice, we report the 
'LEVEL 2' difficulty with AP~(\%) and  APH~(\%) metrics in the main paper.
}
\label{tab:adetails_waymo_centerpoint}
\end{table*}

\begin{table*}[ht!]
\small
\centering

\setlength{\tabcolsep}{4pt}
\begin{tabular}{lll|ccc|ccc|ccc}
\toprule
Backbone & Metric & Method & \multicolumn{3}{c|}{Car} & \multicolumn{3}{c|}{Pedestrian} & \multicolumn{3}{c}{Cyclist} \\
& & & Easy & Mod. & Hard &Easy& Mod. & Hard & Easy& Mod. & Hard\\
\midrule
$R_{40}$ metric & 2D object detection 
& No pre-training     & \textbf{96.86} & 91.78 & 90.88 & {63.31} & 58.76 & {55.63} & 89.39 & 74.71 & 70.30 \\
\cmidrule{3-12}
&& Occupancy-MAE     & 95.65 & {92.12} & \textbf{91.16} & 62.40 & {59.20} & 55.39 & {89.81} & {75.27} & {71.63} \\
\cmidrule{3-12}
&& AD-L-JEPA     & {95.72} & \textbf{93.19} & {91.11} & \textbf{66.65} & \textbf{63.57} & \textbf{60.55} & \textbf{90.92} & \textbf{76.76} & \textbf{73.90} \\
\cmidrule{2-12}
&Bird's eye view 
& No pre-training     & \textbf{93.95} & 87.96 & 86.48 & {54.34} & {50.08} & {45.42} & 83.89 & 67.33 & 63.05 \\
\cmidrule{3-12}
&& Occupancy-MAE     & {93.95} & {88.32} & {87.02} & 51.53 & 48.08 & 43.70 & {85.21} & {68.03} & {63.23} \\
\cmidrule{3-12}
&& AD-L-JEPA     & 92.39 & \textbf{88.54} & \textbf{87.14} & \textbf{58.16} & \textbf{54.05} & \textbf{49.28} & \textbf{86.72} & \textbf{68.51} & \textbf{64.43} \\
\cmidrule{2-12}
&3D object detection 
& No pre-training     & \textbf{90.28} & {79.11} & {76.15} & {48.11} & {44.36} & {39.84} & 78.85 & {62.55} & {58.57} \\
\cmidrule{3-12}
&& Occupancy-MAE     & 88.39 & 79.04 & 75.87 & 47.10 & 43.85 & 39.05 & \textbf{80.87} & \textbf{63.46} & \textbf{59.20} \\
\cmidrule{3-12}
&& AD-L-JEPA     & {88.68} & \textbf{79.48} & \textbf{76.18} & \textbf{52.53} & \textbf{48.48} & \textbf{43.64} & {80.76} & 61.92 & 58.31 \\
\cmidrule{2-12}
&Orientation similarity 
& No pre-training     & \textbf{96.83} & 91.63 & 90.59 & 57.63 & 52.70 & 49.64 & 89.16 & 73.95 & 69.58 \\
\cmidrule{3-12}
&& Occupancy-MAE     & 95.61 & {91.92} & \textbf{90.80} & {58.86} & {54.68} & {50.55} & {89.55} & \textbf{74.53} & {70.90} \\
\cmidrule{3-12}
&& AD-L-JEPA     & {95.69} & \textbf{93.01} & {90.77} & \textbf{62.25} & \textbf{58.51} & \textbf{55.41} & \textbf{90.18} & {74.19} & \textbf{71.24} \\
\midrule\midrule
$R_{11}$ metric & 2D object detection 
& No pre-training     & \textbf{94.48} & 89.36 & 88.43 & {63.24} & 59.72 & {56.33} & 86.10 & {74.33} & 69.72 \\
\cmidrule{3-12}
&& Occupancy-MAE     & 90.63 & {89.50} & \textbf{88.60} & 62.45 & {59.73} & 56.30 & {86.47} & 73.85 & {70.56} \\
\cmidrule{3-12}
&& AD-L-JEPA     & {90.70} & \textbf{89.53} & {88.52} & \textbf{66.13} & \textbf{63.73} & \textbf{60.86} & \textbf{86.98} & \textbf{75.70} & \textbf{72.72} \\
\cmidrule{2-12}
&Bird's eye view 
& No pre-training     & \textbf{89.93} & 87.26 & 83.70 & {54.95} & {51.61} & {47.24} & 82.04 & 66.94 & {62.82} \\
\cmidrule{3-12}
&& Occupancy-MAE     & {89.88} & {87.57} & \textbf{85.23} & 52.12 & 49.54 & 45.81 & {82.16} & {67.83} & 62.57 \\
\cmidrule{3-12}
&& AD-L-JEPA     & 89.84 & \textbf{87.75} & {85.23} & \textbf{58.47} & \textbf{54.76} & \textbf{50.34} & \textbf{83.32} & \textbf{68.16} & \textbf{64.27} \\
\cmidrule{2-12}
&3D object detection 
& No pre-training     & \textbf{88.15} & {77.84} & \textbf{76.15} & {49.69} & {45.78} & {42.34} & 78.22 & {62.46} & \textbf{59.67} \\
\cmidrule{3-12}
&& Occupancy-MAE     & 87.56 & 77.63 & 75.85 & 48.83 & 45.68 & 41.46 & {79.22} & \textbf{63.66} & {59.63} \\
\cmidrule{3-12}
&& AD-L-JEPA     & {87.82} & \textbf{78.11} & {76.09} & \textbf{53.38} & \textbf{49.71} & \textbf{45.56} & \textbf{79.28} & 62.18 & 58.93 \\
\cmidrule{2-12}
&Orientation similarity 
& No pre-training     & \textbf{94.44} & 89.23 & 88.16 & 58.13 & 54.22 & 51.06 & 85.91 & \textbf{73.62} & 69.04 \\
\cmidrule{3-12}
&& Occupancy-MAE     & 90.60 & {89.33} & \textbf{88.28} & {59.19} & {55.67} & {52.01} & {86.26} & 73.21 & {69.85} \\
\cmidrule{3-12}
&& AD-L-JEPA     & {90.68} & \textbf{89.37} & {88.20} & \textbf{62.33} & \textbf{59.25} & \textbf{56.34} & \textbf{86.37} & {73.34} & \textbf{70.31} \\
\bottomrule
\end{tabular}
\caption{Transfer learning experiment: detection results using SECOND on the KITTI3D validation set for all difficulty levels, 
reported with \(R_{40}\) and \(R_{11}\) AP~(\%) across all metrics. The models are pre-trained on Waymo 20\% training data and then fine-tuned using KITTI3D 20\% training data. Following standard practice, we report the 
'3D object detection' metric in the main paper.}
\label{tab:trasfer_kitti_20}
\end{table*}

\begin{table*}[ht!]
\small
\centering

\setlength{\tabcolsep}{4pt}
\begin{tabular}{lll|ccc|ccc|ccc}
\toprule
Backbone & Metric & Method & \multicolumn{3}{c|}{Car} & \multicolumn{3}{c|}{Pedestrian} & \multicolumn{3}{c}{Cyclist} \\
& & & Easy & Mod. & Hard &Easy& Mod. & Hard & Easy& Mod. & Hard\\
\midrule
$R_{40}$ metric & 2D object detection 
& No pre-training     & {95.70} & {94.02} & {91.63} & 66.63 & 61.56 & 58.19 & {89.20} & \textbf{76.78} & {72.30} \\
\cmidrule{3-12}
&& Occupancy-MAE     & 95.54 & 93.68 & 91.59 & {67.86} & {62.67} & {58.89} & \textbf{90.53} & 74.91 & 70.99 \\
\cmidrule{3-12}
&& AD-L-JEPA     & \textbf{97.09} & \textbf{94.35} & \textbf{91.66} & \textbf{70.15} & \textbf{66.35} & \textbf{62.80} & 88.36 & {76.71} & \textbf{73.49} \\
\cmidrule{2-12}
&Bird's eye view 
& No pre-training     & \textbf{93.33} & {89.40} & {87.30} & \textbf{59.45} & {53.36} & {48.50} & 84.15 & {67.93} & {63.62} \\
\cmidrule{3-12}
&& Occupancy-MAE     & 92.00 & 88.02 & 87.07 & {58.73} & 52.87 & 47.66 & \textbf{87.29} & 67.49 & 63.29 \\
\cmidrule{3-12}
&& AD-L-JEPA     & {92.30} & \textbf{89.96} & \textbf{87.39} & 58.07 & \textbf{54.34} & \textbf{49.58} & {84.89} & \textbf{69.01} & \textbf{65.01} \\
\cmidrule{2-12}
&3D object detection 
& No pre-training     & {89.91} & 81.05 & 78.12 & {54.19} & {48.75} & {43.74} & {81.30} & 62.83 & 59.10 \\
\cmidrule{3-12}
&& Occupancy-MAE     & 89.87 & {81.20} & {78.22} & 53.70 & 48.17 & 42.88 & \textbf{84.48} & {64.09} & {59.88} \\
\cmidrule{3-12}
&& AD-L-JEPA     & \textbf{90.59} & \textbf{81.55} & \textbf{78.43} & \textbf{54.37} & \textbf{50.13} & \textbf{45.58} & 81.25 & \textbf{64.12} & \textbf{60.05} \\
\cmidrule{2-12}
&Orientation similarity 
& No pre-training     & {95.67} & {93.87} & \textbf{91.39} & 61.96 & 56.37 & 52.80 & {88.95} & {75.92} & {71.46} \\
\cmidrule{3-12}
&& Occupancy-MAE     & 95.52 & 93.53 & 91.32 & {62.98} & {57.03} & {53.17} & \textbf{90.32} & 74.40 & 70.43 \\
\cmidrule{3-12}
&& AD-L-JEPA     & \textbf{97.07} & \textbf{94.16} & {91.35} & \textbf{64.07} & \textbf{60.06} & \textbf{56.28} & 88.16 & \textbf{75.98} & \textbf{72.77} \\
\midrule\midrule
$R_{11}$ metric & 2D object detection 
& No pre-training     & {90.81} & {89.78} & 88.93 & 66.43 & 61.74 & 58.81 & 85.53 & \textbf{75.43} & {71.28} \\
\cmidrule{3-12}
&& Occupancy-MAE     & 90.78 & 89.77 & {88.94} & {67.18} & {62.62} & {59.50} & \textbf{87.10} & 74.23 & 70.53 \\
\cmidrule{3-12}
&& AD-L-JEPA     & \textbf{94.95} & \textbf{89.85} & \textbf{88.96} & \textbf{69.76} & \textbf{66.21} & \textbf{62.59} & {85.78} & {75.37} & \textbf{72.19} \\
\cmidrule{2-12}
&Bird's eye view 
& No pre-training     & {89.76} & {87.58} & {86.10} & \textbf{60.22} & {54.06} & {49.88} & 81.41 & 67.30 & {63.49} \\
\cmidrule{3-12}
&& Occupancy-MAE     & 89.62 & 87.39 & 85.52 & {59.17} & 53.50 & 48.89 & \textbf{84.68} & {67.68} & 62.92 \\
\cmidrule{3-12}
&& AD-L-JEPA     & \textbf{89.84} & \textbf{87.65} & \textbf{86.25} & 58.13 & \textbf{55.19} & \textbf{50.48} & {82.47} & \textbf{68.48} & \textbf{64.82} \\
\cmidrule{2-12}
&3D object detection 
& No pre-training     & 87.90 & 78.07 & 76.80 & \textbf{54.73} & {50.53} & {45.54} & {79.91} & 62.95 & 58.86 \\
\cmidrule{3-12}
&& Occupancy-MAE     & {88.08} & {78.27} & {76.97} & 54.35 & 49.67 & 45.04 & \textbf{82.89} & \textbf{63.80} & {60.14} \\
\cmidrule{3-12}
&& AD-L-JEPA     & \textbf{88.43} & \textbf{78.38} & \textbf{77.01} & {54.70} & \textbf{52.04} & \textbf{47.21} & 79.47 & {63.53} & \textbf{60.42} \\
\cmidrule{2-12}
&Orientation similarity 
& No pre-training     & {90.79} & {89.66} & \textbf{88.72} & 62.34 & 57.25 & 54.09 & 85.33 & {74.68} & {70.52} \\
\cmidrule{3-12}
&& Occupancy-MAE     & 90.77 & 89.65 & {88.71} & {62.92} & {57.74} & {54.51} & \textbf{86.94} & 73.77 & 70.03 \\
\cmidrule{3-12}
&& AD-L-JEPA     & \textbf{94.92} & \textbf{89.69} & 88.69 & \textbf{64.35} & \textbf{60.85} & \textbf{56.99} & {85.59} & \textbf{74.74} & \textbf{71.53} \\
\bottomrule
\end{tabular}

\caption{Transfer learning experiment: detection results using SECOND on the KITTI3D validation set for all difficulty levels, 
reported with \(R_{40}\) and \(R_{11}\) AP~(\%) across all metrics. The models are pre-trained on Waymo 20\% training data and then fine-tuned using KITTI3D 50\% training data. Following standard practice, we report the 
'3D object detection' metric in the main paper.}
\label{tab:trasfer_kitti_50}
\end{table*}

\begin{table*}[ht!]
\small
\centering
\setlength{\tabcolsep}{4pt}
\resizebox{0.8\textwidth}{!}{
\begin{tabular}{lll|ccc|ccc|ccc}
\toprule
Backbone & Metric & Method & \multicolumn{3}{c|}{Car} & \multicolumn{3}{c|}{Pedestrian} & \multicolumn{3}{c}{Cyclist} \\
& & & Easy & Mod. & Hard &Easy& Mod. & Hard & Easy& Mod. & Hard\\
\midrule
$R_{40}$ metric & 2D object detection 
& No pre-training     & 96.29 & 94.05 & 91.63 & 68.61 & 65.73 & 62.17 & 89.06 & 77.96 & 73.99 \\
\cmidrule{3-12}
&& Occupancy-MAE     & \textbf{97.35} & {94.56} & 91.89 & 69.25 & 65.46 & 62.23 & 92.19 & {79.89} & 76.35 \\
\cmidrule{3-12}
&& Occupancy-MAE\textsuperscript{\textdagger}     & 95.56 & 94.23 & 91.80 & 69.28 & 66.32 & 62.54 & 89.36 & 78.35 & 74.70 \\
\cmidrule{3-12}
&& AD-L-JEPA     & {97.15} & \textbf{94.63} & {91.98} & \textbf{71.13} & \textbf{67.94} & \textbf{64.83} & {92.78} & 79.85 & {76.85} \\
\cmidrule{3-12}
&& AD-L-JEPA\textsuperscript{\textdagger}     & 95.56 & 94.38 & \textbf{93.07} & {69.40} & {66.50} & {62.94} & \textbf{93.07} & \textbf{80.19} & \textbf{77.05} \\
\cmidrule{2-12}
&Bird's eye view 
& No pre-training     & {93.27} & \textbf{90.28} & \textbf{87.85} & \textbf{63.37} & \textbf{58.30} & {53.94} & 85.58 & 70.49 & 66.03 \\
\cmidrule{3-12}
&& Occupancy-MAE     & 92.24 & 89.82 & 87.47 & 61.08 & 56.12 & 51.98 & 86.77 & 71.00 & 66.54 \\
\cmidrule{3-12}
&& Occupancy-MAE\textsuperscript{\textdagger}     & \textbf{93.62} & 89.96 & 87.70 & 59.36 & 54.36 & 50.34 & 86.99 & {71.44} & {67.36} \\
\cmidrule{3-12}
&& AD-L-JEPA     & 92.47 & {90.23} & {87.71} & {62.14} & {58.01} & \textbf{54.21} & \textbf{88.98} & 71.12 & 66.92 \\
\cmidrule{3-12}
&& AD-L-JEPA\textsuperscript{\textdagger}     & 91.71 & 89.57 & 87.40 & 60.92 & 56.88 & 52.71 & {87.83} & \textbf{72.47} & \textbf{68.14} \\
\cmidrule{2-12}
&3D object detection 
& No pre-training     & 90.46 & \textbf{81.99} & \textbf{79.05} & \textbf{57.12} & 52.02 & 47.53 & 80.36 & 65.07 & 60.99 \\
\cmidrule{3-12}
&& Occupancy-MAE     & 90.37 & 81.65 & 78.66 & 56.69 & 51.51 & 46.95 & 83.77 & 66.72 & 62.53 \\
\cmidrule{3-12}
&& Occupancy-MAE\textsuperscript{\textdagger}     & \textbf{90.73} & 81.78 & {78.77} & 53.79 & 48.92 & 44.35 & 84.06 & {67.34} & {63.55} \\
\cmidrule{3-12}
&& AD-L-JEPA     & {90.72} & {81.83} & 78.62 & 56.95 & {52.41} & \textbf{48.32} & {84.68} & 66.73 & 62.47 \\
\cmidrule{3-12}
&& AD-L-JEPA\textsuperscript{\textdagger}     & 89.29 & 80.92 & 78.18 & {57.00} & \textbf{52.45} & {48.05} & \textbf{85.03} & \textbf{69.76} & \textbf{65.49} \\
\cmidrule{2-12}
&Orientation similarity 
& No pre-training     & 96.28 & 93.92 & 91.42 & 63.41 & 60.21 & 56.35 & 88.68 & 76.90 & 72.93 \\
\cmidrule{3-12}
&& Occupancy-MAE     & \textbf{97.31} & {94.35} & 91.58 & {65.47} & {60.93} & {57.31} & 92.05 & 79.38 & 75.81 \\
\cmidrule{3-12}
&& Occupancy-MAE\textsuperscript{\textdagger}     & 95.54 & 94.10 & 91.60 & 63.28 & 59.90 & 55.84 & 89.11 & 77.70 & 74.03 \\
\cmidrule{3-12}
&& AD-L-JEPA     & {97.13} & \textbf{94.49} & {91.73} & \textbf{65.53} & \textbf{61.85} & \textbf{58.35} & {92.62} & {79.44} & {76.37} \\
\cmidrule{3-12}
&& AD-L-JEPA\textsuperscript{\textdagger}     & 95.55 & 94.24 & \textbf{92.84} & 64.09 & 60.60 & 56.80 & \textbf{92.94} & \textbf{79.74} & \textbf{76.54} \\
\midrule\midrule
$R_{11}$ metric & 2D object detection 
& No pre-training     & 94.10 & 89.88 & 89.03 & 68.18 & 65.61 & 62.15 & 85.94 & 76.27 & 72.67 \\
\cmidrule{3-12}
&& Occupancy-MAE     & {94.91} & {90.01} & 89.19 & 68.56 & 65.05 & 62.40 & {89.85} & {78.78} & 75.17 \\
\cmidrule{3-12}
&& Occupancy-MAE\textsuperscript{\textdagger}     & 90.76 & 89.90 & 89.13 & {68.87} & 65.89 & 63.15 & 86.01 & 77.00 & 73.83 \\
\cmidrule{3-12}
&& AD-L-JEPA     & \textbf{94.92} & \textbf{90.06} & \textbf{89.24} & \textbf{70.22} & \textbf{67.47} & \textbf{64.87} & 89.47 & 78.25 & \textbf{75.71} \\
\cmidrule{3-12}
&& AD-L-JEPA\textsuperscript{\textdagger}     & 90.73 & 89.91 & {89.20} & 68.58 & {66.18} & {63.16} & \textbf{91.81} & \textbf{79.01} & {75.32} \\
\cmidrule{2-12}
&Bird's eye view 
& No pre-training     & \textbf{90.02} & \textbf{88.20} & \textbf{86.81} & \textbf{63.78} & \textbf{58.73} & \textbf{55.46} & 83.26 & 69.84 & 65.82 \\
\cmidrule{3-12}
&& Occupancy-MAE     & 89.76 & 87.81 & 86.28 & 61.37 & 56.18 & 53.01 & 83.04 & 70.47 & 65.30 \\
\cmidrule{3-12}
&& Occupancy-MAE\textsuperscript{\textdagger}     & {89.96} & 87.95 & {86.75} & 60.11 & 54.68 & 51.27 & 83.98 & {70.48} & {67.06} \\
\cmidrule{3-12}
&& AD-L-JEPA     & 89.91 & {88.09} & 86.69 & {62.69} & {58.05} & {55.21} & \textbf{87.61} & 70.06 & 66.13 \\
\cmidrule{3-12}
&& AD-L-JEPA\textsuperscript{\textdagger}     & 89.35 & 87.59 & 86.42 & 61.51 & 57.36 & 54.07 & {84.01} & \textbf{71.31} & \textbf{67.58} \\
\cmidrule{2-12}
&3D object detection 
& No pre-training     & 88.47 & \textbf{78.89} & \textbf{77.69} & {57.33} & {53.78} & 48.79 & 79.79 & 64.93 & 60.55 \\
\cmidrule{3-12}
&& Occupancy-MAE     & 88.29 & 78.47 & 77.18 & 56.17 & 52.60 & 48.10 & 80.47 & {67.23} & 62.01 \\
\cmidrule{3-12}
&& Occupancy-MAE\textsuperscript{\textdagger}     & \textbf{88.65} & 78.61 & {77.34} & 53.73 & 49.49 & 46.10 & {81.78} & 66.84 & {63.49} \\
\cmidrule{3-12}
&& AD-L-JEPA     & {88.61} & {78.65} & 77.20 & 56.98 & 53.62 & \textbf{49.53} & 81.55 & 67.12 & 61.77 \\
\cmidrule{3-12}
&& AD-L-JEPA\textsuperscript{\textdagger}     & 87.42 & 77.90 & 76.86 & \textbf{57.37} & \textbf{54.01} & {49.44} & \textbf{82.87} & \textbf{69.30} & \textbf{64.74} \\
\cmidrule{2-12}
&Orientation similarity 
& No pre-training     & 94.08 & 89.77 & 88.85 & 63.63 & 60.70 & 56.97 & 85.59 & 75.27 & 71.73 \\
\cmidrule{3-12}
&& Occupancy-MAE     & {94.87} & {89.83} & 88.91 & {65.14} & {61.09} & {58.00} & {89.68} & {78.31} & 74.66 \\
\cmidrule{3-12}
&& Occupancy-MAE\textsuperscript{\textdagger}     & 90.75 & 89.80 & 88.96 & 63.58 & 60.33 & 57.32 & 85.82 & 76.37 & 73.19 \\
\cmidrule{3-12}
&& AD-L-JEPA     & \textbf{94.90} & \textbf{89.95} & \textbf{89.01} & \textbf{65.17} & \textbf{62.13} & \textbf{59.16} & 89.32 & 77.89 & \textbf{75.23} \\
\cmidrule{3-12}
&& AD-L-JEPA\textsuperscript{\textdagger}     & 90.72 & 89.80 & {89.01} & 63.90 & 61.06 & 57.83 & \textbf{91.66} & \textbf{78.57} & {74.85} \\
\bottomrule
\end{tabular}
}
\caption{Transfer learning experiment: detection results using SECOND on the KITTI3D validation set for all difficulty levels, 
reported with \(R_{40}\) and \(R_{11}\) AP~(\%) across all metrics. The models are pre-trained on Waymo 20\% training data and then fine-tuned using KITTI3D 100\% training data.  \textdagger\ denotes that the first layer of the model is randomly initialized. Following standard practice, we report the 
'3D object detection' metric in the main paper.}
\label{tab:trasfer_kitti_100}
\end{table*}

\begin{table*}[ht]
\small
\centering
\setlength{\tabcolsep}{4pt}
\begin{tabular}{l|cccc|cccc|cccc|c}
\toprule
\textbf{Method}
  & \multicolumn{4}{c|}{\textbf{Vehicle}}
  & \multicolumn{4}{c|}{\textbf{Pedestrian}}
  & \multicolumn{4}{c|}{\textbf{Cyclist}}
  & \textbf{mAP} \\
& \textbf{overall} & \textbf{0--30} & \textbf{30--50} & \textbf{50--inf}
& \textbf{overall} & \textbf{0--30} & \textbf{30--50} & \textbf{50--inf}
& \textbf{overall} & \textbf{0--30} & \textbf{30--50} & \textbf{50--inf}
& \\
\midrule
baseline (SECOND)
  & 71.19 & 84.04 & 63.02 & 47.25
  & 26.44 & 29.33 & 24.05 & 18.05
  & 58.04 & 69.96 & 52.43 & 34.61
  & 51.89 \\
\midrule
\multicolumn{14}{l}{\textit{100k pre-training data}} \\
BYOL
  & 68.02 & 81.01 & 60.21 & 44.17
  & 19.50 & 22.16 & 16.68 & 12.06
  & 50.61 & 62.46 & 44.29 & 28.18
  & 46.04 \\
PointContrast
  & 71.07 & 83.31 & 64.90 & 49.34
  & 22.52 & 23.73 & 21.81 & 16.06
  & 56.36 & 68.11 & 50.35 & 34.06
  & 49.98 \\
SwAV
  & 72.71 & 83.68 & 65.91 & 50.10
  & 25.13 & 27.77 & 22.77 & 16.36
  & 58.05 & 69.99 & 52.23 & 34.86
  & 51.96 \\
DeepCluster
  & 73.19 & 84.25 & 66.86 & 50.47
  & 24.00 & 26.36 & 21.73 & 16.79
  & \textbf{58.99} & \textbf{70.80} & \textbf{53.66} & \textbf{36.17}
  & 52.06 \\
DepthContrast
  & 71.88 & 84.26 & 65.58 & 49.97
  & 23.57 & 26.36 & 21.15 & 14.39
  & 56.63 & 68.26 & 50.82 & 34.67
  & 50.69 \\
ProposalContrast
  & 72.99 & \textbf{84.41} & 65.92 & 50.11
  & 25.77 & 27.95 & 23.74 & 18.06
  & 58.23 & 69.99 & 53.03 & 35.48
  & 52.33 \\
ALSO
  & 71.73 & 84.30 & 65.21 & 48.30
  & 28.16 & 31.45 & 25.19 & 16.29
  & 58.13 & 70.04 & 52.76 & 33.88
  & 52.68 \\
Occupancy-MAE
  &  \textbf{73.54} & 84.15 & \textbf{68.08}  & \textbf{51.42} 
  &  25.93 & 28.94 &22.98   & 15.79  
  &  58.34 & 70.34 & 52.13  & 35.73
  &  52.60\\
AD-L-JEPA
  &  73.18 &84.29  &66.18  & 49.63
  &  \textbf{29.19} &\textbf{32.66}  &\textbf{25.61}  & \textbf{18.56}
  &  58.14 &70.20  &52.53  &34.24
  &  \textbf{53.50}\\
\midrule
\multicolumn{14}{l}{\textit{500k pre-training data}} \\
BYOL
  & 70.93 & 84.15 & 63.48 & 45.74
  & 25.86 & 29.91 & 21.55 & 15.83
  & 55.63 & 58.59 & 49.01 & 29.53
  & 50.82 \\
PointContrast
  & 71.39 & 83.89 & 65.22 & 47.73
  & 27.69 & 32.53 & 23.00 & 14.68
  & 56.88 & 69.01 & 50.41 & 34.57
  & 51.99 \\
SwAV
  & 72.51 & 83.39 & 65.46 & \textbf{51.08}
  & 27.08 & 29.94 & 25.19 & 17.13
  & 57.85 & 69.87 & 52.38 & 33.78
  & 52.48 \\
DeepCluster
  & 71.62 & 83.99 & 65.55 & 50.77
  & 29.33 & 33.25 & 25.08 & 17.00
  & 57.61 & 68.57 & 52.58 & 34.05
  & 52.86 \\
DepthContrast
  & 71.92 & 84.38 & 65.86 & 48.48
  & 29.01 & 33.09 & 24.23 & 15.88
  & 57.51 & 69.86 & 51.00 & 34.41
  & 52.81 \\
AD-L-JEPA
  & \textbf{73.25} &\textbf{85.30}  & \textbf{67.46} &51.04
  & \textbf{31.91} &\textbf{35.83}  & \textbf{28.25} &\textbf{17.61} 
  & \textbf{59.47} &\textbf{71.08}  & \textbf{53.76} &\textbf{37.35}
  & \textbf{54.87} \\
\midrule
\multicolumn{14}{l}{\textit{1M pre-training data}} \\
BYOL
  & 71.32 & 83.59 & 64.89 & 50.27
  & 25.02 & 27.06 & 22.96 & 17.04
  & 58.56 & 70.18 & 52.74 & 36.32
  & 51.63 \\
PointContrast
  & 71.87 & \textbf{86.93} & 62.85 & 48.65
  & 28.03 & 33.07 & 25.91 & 14.44
  & \textbf{60.88} & 71.12 & \textbf{55.77} & \textbf{36.78}
  & 53.59 \\
SwAV
  & 72.46 & 83.09 & 66.66 & 51.50
  & 29.84 & 34.15 & 26.22 & 17.61
  & 57.84 & 68.79 & 52.21 & 35.39
  & 53.38 \\
DeepCluster
  & 72.89 & 83.52 & \textbf{67.09} & 50.38
  & 30.32 & 34.76 & 26.43 & \textbf{18.33}
  & 57.94 & 69.18 & 52.42 & 34.36
  & 53.72 \\
AD-L-JEPA
  & \textbf{73.01} & 83.89 & 66.75 & \textbf{52.02} 
  & \textbf{31.94} & \textbf{35.83}& \textbf{28.10} & 17.81
  & 59.16 & \textbf{71.21} & 53.01 & 36.63
  & \textbf{54.70} \\
\bottomrule
\end{tabular}
\caption{Detection performance on ONCE val split. Methods are grouped by pre-training set size (100k, 500k, 1M). We report AP (\%) for Vehicle, Pedestrian, Cyclist at three distance ranges (meters) and their overall mAP.}
\label{tab:once}
\end{table*}

\end{document}